\documentclass[11pt]{article}

\usepackage[final]{acl} 

\usepackage{times}
\usepackage{latexsym}

\usepackage[T1]{fontenc}

\usepackage[utf8]{inputenc}

\usepackage{microtype}

\usepackage{inconsolata}

\usepackage{graphicx}

\usepackage{float}

\usepackage{amsmath,amssymb,booktabs,enumitem}
\usepackage{graphicx}
\usepackage{xcolor}
\usepackage{hyperref}
\usepackage{natbib}

\title{The Emergence of Relevance Through Axiomatic \\ Attention Patterns During LoRA Fine-Tuning}

\author{
    Matthew Perlman \quad Atharva Nijasure \quad James Allan \\
    University of Massachusetts Amherst \\
    \texttt{\{mperlman, anijasure, allan\}@umass.edu}
}

\begin{document}

\maketitle
\begin{abstract}
LoRA fine-tuning is standard for adapting LLMs to reranking, but it remains unclear where in the network task-specific relevance behavior is learned and what attention-level changes accompany that learning. Through ablation and attention experiments, we identify where LoRA attention updates to RankLLaMA improve performance and whether those gains coincide with interpretable relevance-oriented attention patterns such as lexical matching, rarity sensitivity, and query-document interaction. We find that given LoRA fine-tuned MLPs throughout the network, restricting LoRA attention updates to a compact mid-network region is sufficient for recovering over half of the performance gained by applying LoRA to all attention layers, and that omitting attention fine-tuning in this region hurts performance more than elsewhere in the network. Additionally, we show that regions where applying LoRA affects performance the most overlap with regions where fine-tuning increased attention to axiomatic IR features. Rarity sensitivity, document-query interaction, and several compositional features are highly correlated with gains in ranking performance. Our results support an interpretable, correlational account of how relevance-oriented behavior emerges during LoRA fine-tuning and point toward improved strategies for adapting rerankers.
\end{abstract}

\section{Introduction}

Neural rerankers built on large language models now represent the state of the art in multi-stage retrieval pipelines \citep{ma2024rankllama}.
Among the techniques used to adapt these models to ranking, Low-Rank Adaptation (LoRA) \citep{hu2022lora} has become a practical default.
It substantially reduces the number of trainable parameters while often recovering the effectiveness of full fine-tuning in many language modeling tasks.
While this makes adaptation efficient in practice, it also raises fundamental interpretability questions: what relevance signals are learned during fine-tuning, where in the network is this behavior implemented, and does it align with classical information retrieval signals?

These questions are especially natural in reranking, because the field of information retrieval (IR) has historically grounded relevance in principled axiomatic signals such as lexical matching, rarity sensitivity, term frequency weighting, and query-document interaction patterns \citep{fang2005axiomatic, robertson2009bm25}. A growing body of work has begun to recover such signals within neural rerankers \citep{lu2025pathway, tanya2025probing}, but whether and where axiomatic IR features are learned during LoRA fine-tuning, and whether the emergence of these features improves performance, is unknown.

We study these questions through two complementary experiments and a joint analysis of their results. First, we perform a systematic set of head-wise, layer-wise, and window-wise ablations to localize where LoRA updates to the transformer's attention matrices are most critical for reranking performance.
Second, we quantify how attention mass shifts over interpretable token-pairing categories associated with axiomatic IR properties, providing a fine-grained view of relevance-oriented attention patterns that emerge during fine-tuning.
Finally, we combine these analyses to identify overlap between the regions where LoRA updates are performance-critical and the regions where fine-tuning learns interpretable IR-aligned attention patterns, providing evidence that LoRA installs ranking behavior in a structured and interpretable manner.

\section{Related Work}

\paragraph{Neural Reranking and LoRA.}
The use of pre-trained transformers for passage reranking, initially through cross-encoders, has substantially advanced the field \citep{nogueira2019passage}.
\citet{ma2024rankllama} establish the RankLLaMA model we study: a decoder-only LLM fine-tuned with LoRA for pointwise relevance scoring.

\paragraph{Mechanistic Interpretability of Rerankers.}
A growing body of work examines what neural rerankers learn internally.
\citet{lu2025pathway} show that a cross-encoder can implement a semantic variant of BM25: specific attention heads compute soft term frequency and saturation effects, while a low-rank subspace of the embedding matrix encodes IDF-like signals.
\citet{tanya2025probing} probe \textit{MLP} activations of fine-tuned reranking LLMs layer by layer, finding that classical IR features such as query term coverage and term frequency are prominently encoded.
Both works demonstrate that neural rerankers can recover interpretable IR-style relevance behavior, but neither addresses \textit{how} these behaviors are acquired during fine-tuning or which attention components are responsible.
Our work is complementary: we focus on the attention mechanism and use structured ablations to directly study what LoRA fine-tuning changes.

\paragraph{LoRA Ablations in IR.}
\citet{nijasure2025lora} study LoRA rank, checkpoint trajectories, and MLP vs.\ MHA contributions across multiple fine-tuned reranking LLMs, finding that MHA updates offer a stronger boost than MLP-only updates.
Their ablations, however, consider only coarse all-attention or all-MLP conditions.
In contrast, this paper performs MHA ablations at the head, layer, and window level, enabling a substantially more localized account of where LoRA attention updates are functionally important.

\paragraph{Attention Head Specialization.}
Beyond the reranking setting, a rich literature has characterized how individual attention heads develop interpretable specialized functions. \citet{voita2019analyzing} demonstrate that a small subset of heads accounts for most model performance in neural machine translation, playing identifiable roles: attending to positional neighbors, tracking syntactic dependencies, or routing attention toward \textit{rare tokens}. This last function directly parallels the rarity-sensitive attention patterns we study. \citet{wang2023interpretability} introduce path patching and use it to identify a 26-head circuit in GPT-2 small responsible for indirect object identification, demonstrating that discrete linguistic behaviors are implemented by specific multi-head compositions. Their ablation methodology of selectively restoring or removing components and measuring task performance directly informs our keep/omit design. \citet{olsson2022context} further show that \textit{induction heads}, a two-layer match-and-copy circuit, emerge during training and are causally responsible for in-context learning, providing early evidence that behavioral transitions during training can be localized to specific attention components.

More recent work has extended these ideas to fine-tuning dynamics. \citet{wu2024finetuning} show that instruction tuning shifts attention patterns in LLMs, causing the model to preferentially attend to instruction tokens --- evidence that the adaptation process systematically reconfigures how attention is distributed across the network. Work on post-training of reasoning models finds that fine-tuning induces newly emergent attention head circuits absent from the base model, isolable through ablation \citep{park2025sparks}. These results support a general principle: fine-tuning concentrates meaningful attention changes in a sparse, task-relevant subset of components, a principle our work extends to the context of LoRA-based relevance adaptation.

\paragraph{Axiomatic IR.}
Axiomatic IR properties serve as a formalization of what well-behaved retrieval models should satisfy, including term frequency weighting, rare word sensitivity, and document-query interaction \citep{fang2005axiomatic, robertson2009bm25}. \citet{chen2024axiomatic} apply these axioms to causal interventions for reverse-engineering neural retrieval models. The token-pair features we study draw directly from such axioms, interpreting attention behavior through established relevance principles.

\section{Research Questions}

\begin{enumerate}[label=\textbf{RQ\arabic*.}, leftmargin=*, itemsep=4pt]
  \item Where does LoRA fine-tuning attention \\ contribute most to reranking performance?
  \item Does LoRA fine-tuning teach attention heads to attend to axiomatic IR patterns?
  \item Do regions where LoRA updates improve performance coincide with regions where axiomatic attention patterns emerge?
\end{enumerate}

\section{Ablation Experiments}

The first experimental question is where LoRA attention updates contribute most to reranking performance (RQ1). To isolate which components drive performance, we compare the fully LoRA fine-tuned RankLLaMA, the base model of Rank-LLaMA, and a suite of RankLLaMA variants with targeted attention matrix ablations.

\subsection{Methodology}

\paragraph{Models.} Our experiments use RankLLaMA-7B, a decoder-only reranker adapted using LoRA from \textit{meta-llama/Llama-2-7b-hf} for pointwise relevance scoring following \citet{ma2024rankllama}. 
The backbone 
consists of 32 transformer layers, each containing 32 attention heads with a total
hidden dimension of 4096. LoRA adapters (rank $r = 32$, $\alpha = 64$) are 
injected into the attention weight matrices and the MLP of every layer, leaving the base model frozen. \citet{ma2024rankllama} release separate checkpoints trained for passage and document reranking; all results in the main paper refer to the \textit{passage} variant. We report a replication of our main findings on the \textit{document} variant in \S\ref{app:generalization}.

\paragraph{Ablation Strategies.} Ablations are performed only on the attention matrices; the MLP receives LoRA updates regardless of ablation condition. This ensures observed performance differences are attributable to attention components alone, isolating the cause of interpretable changes. Our analysis comprises two complementary ablation strategies that differ in how LoRA updates are applied with respect to the component under investigation:

\begin{itemize}
    \item \textbf{Keep} ablations apply LoRA attention updates only to the component under investigation. All other attention weight matrices revert to their \textit{base model parameters} — that is, the pre-fine-tuning weights of the original model.
    
    \item \textbf{Omit} ablations apply LoRA attention updates to the full network \textit{except} the component under investigation, reverting the attention weight matrices of heads in the given component back to their base model parameters.
\end{itemize}

Keep ablations reveal which updates are \textit{sufficient} for good performance; omit ablations reveal which are \textit{necessary} by measuring the cost of their exclusion. Together, these strategies provide complementary views of how performance is affected by fine-tuning a given component.
The choice to revert to base model parameters rather than zero or mean ablating ensures that performance differences reflect the \textit{value of fine-tuning} a given component, not the \textit{cost of destroying} it.

\paragraph{Component Granularity.} We perform ablations at three granularities: head, layer, and window.

\begin{itemize}
    \item \textbf{Single Head ablations} consider the effect of applying LoRA to individual attention heads.
    These ablations provide the finest resolution, revealing whether a given head learns critical relevance behavior during fine-tuning.
    
    \item \textbf{Single Layer ablations} consider the effect of applying LoRA to a single attention layer.
    These ablations aggregate the presence of performance-critical behavior across all heads within a layer, smoothing head-level variation to reveal broader layer-wise trends in where fine-tuning matters.
        
    \item \textbf{Window ablations} consider the effect of LoRA application to a contiguous window of layers.
    These ablations are motivated by the transformer circuits hypothesis \citep{elhage2021framework}, that coherent functional behaviors may emerge from the coordinated interaction of consecutive layers rather than isolated layers.
\end{itemize}

Together, the three granularities form a hierarchical view of where LoRA fine-tuning learns relevance-oriented behavior. Single head ablations reveal fine-grained 
specialization, layer-level ablations identify where critical heads are concentrated, 
and window ablations determine whether a group of layers reflects a coherent, circuit-like region.

\paragraph{Evaluation Metrics.}
We evaluate each ablation over the same 50 queries randomly sampled from the MS MARCO Dev split \citep{bajaj2016msmarco}. The number of candidate documents per query differs by ablation granularity to balance computational cost against metric sensitivity. Per-head ablations use 10 candidate documents per query (one relevant and nine random non-relevant) while layer- and window-wise ablations use 100 candidate documents per query (also one relevant, but 99 random non-relevant). The candidate sets are the same for each set of 10 or 100 documents. Each model variant reranks the full candidate set for every query, and performance is measured over the resulting rankings and score distributions. These candidate sets are intentionally small, and we sample random negatives rather than intentionally hard candidates. Our goal is to detect relative behavioral differences across ablations rather than to measure absolute retrieval performance.

Single head ablations are often too minimal to affect the ranking of the relevant document. To account for this, we incorporate two evaluation metrics: one rank-based and one score-based.

\begin{itemize}
    \item \textbf{NDCG} \citep{jarvelin2002ndcg} measures ranking quality by rewarding placements of the relevant document higher in the ranked list while discounting lower-ranked positions logarithmically. As an industry standard, NDCG serves as our primary ranking metric for evaluating whether an ablation meaningfully changes retrieval effectiveness.

    \item \textbf{Mean Score Margin} is defined as the mean difference between the score assigned to the relevant document and the scores assigned to each non-relevant document. Mean score margin provides a continuous measure of relevance separation and remains sensitive even when the ranking itself is unchanged. This is particularly important for granular ablations, where small perturbations often shift the score distribution without affecting the final ordering of the ranked list.
\end{itemize}

\begin{table}[h!]
\centering
\small
\begin{tabular}{lcc}
\toprule
 & \textbf{NDCG} & \textbf{Mean Score Margin} \\
\midrule
\textbf{Base Model} & 0.199 & $-$0.204 \\
\textbf{Fine-Tuned Model} & 0.911 & 8.768 \\
\midrule
\textbf{Performance Gain} & 0.712 & 8.972 \\
\bottomrule
\end{tabular}
\caption{Performance of the base model and fine-tuned model evaluated over 50 queries $\times$ 100 documents.}
\label{tab:performance}
\end{table}

Table~\ref{tab:performance} shows NDCG and Mean Score Margin performance for the base and fine-tuned models.
We calculate the performance of the base and fine-tuned models separately for single head ablations to reflect the smaller candidate set of 10 (See \S\ref{app:control}).

Omit ablations reflect the performance lost by omitting a component from full LoRA fine-tuning, so we report the signed performance difference between the ablated model performance and the fully LoRA fine-tuned model.
For keep ablations, we report the signed performance difference between the ablated and base model, representing the gain of fine-tuning only the given component and MLPs.

\subsection{Results}


\begin{figure}[t!]
    \includegraphics[width=\linewidth]{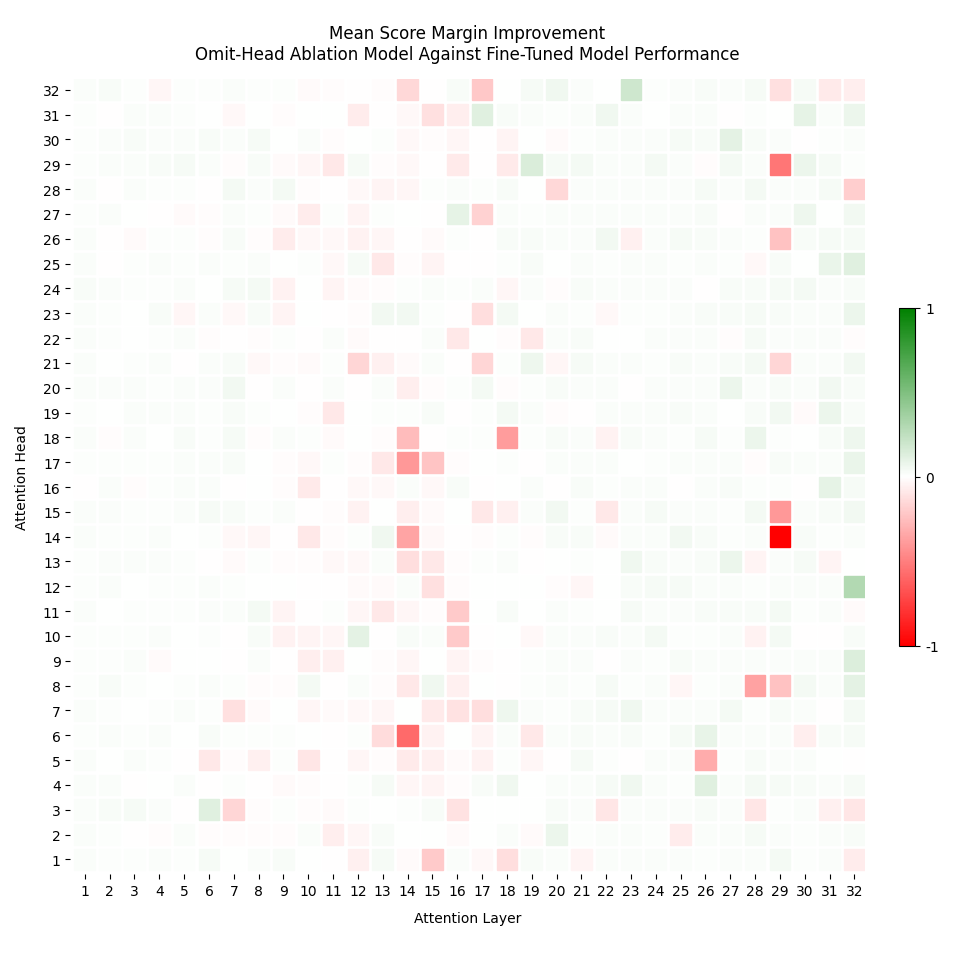}
    \caption{The performance difference between the fine-tuned model and ablated model, measured per-{\bf omitted-head} in terms of mean score margin. Scores are normalized by the maximum absolute value across all heads, such that the scale reflects relative importance in the network.
    Red indicates omitting the given head hurts performance and green indicates it improves performance.
    Omitting fine-tuning on many of the heads in layers 7 through 19 and 26 through 29 hurts performance most. Several heads in layers 19 through 32 perform marginally better using the base model parameters.}
    \label{fig:omit-head}
\end{figure}

\begin{figure}[t!]
    \includegraphics[width=\linewidth]{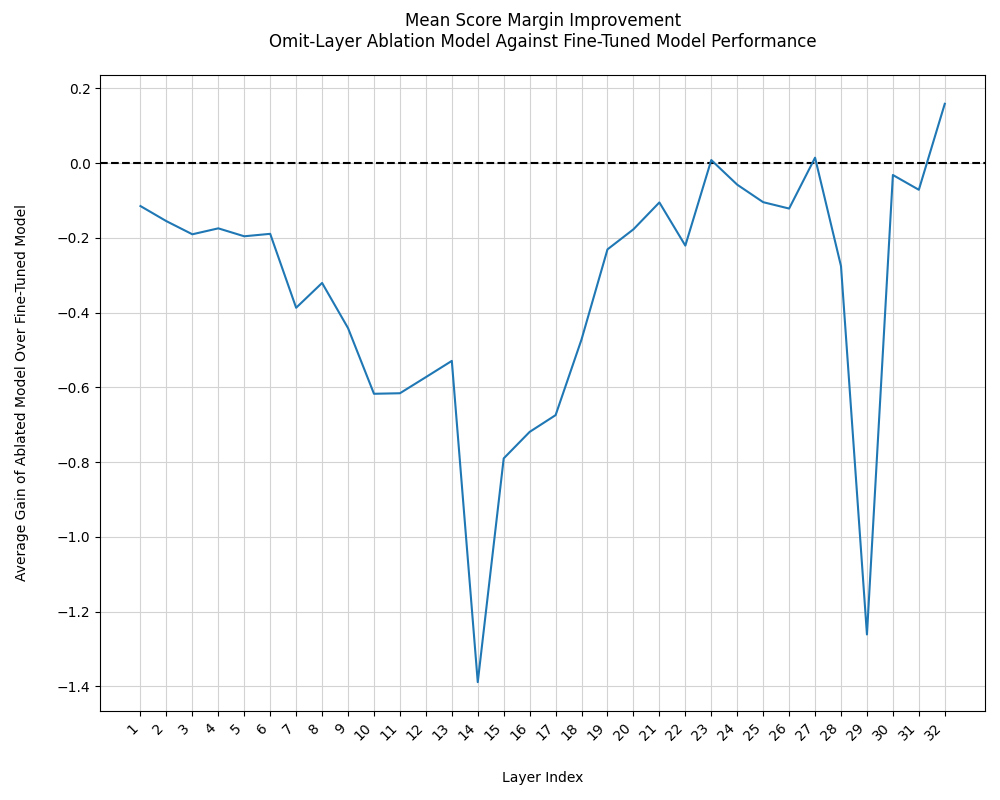}
    \caption{The performance difference between the fine-tuned model and ablated model, measured per-{\bf omitted-layer} in terms of mean score margin. Omitting layers 14 and 29 hurts performance most.}
    \label{fig:omit-layer}
\end{figure}

\begin{figure}[t]
    \includegraphics[width=\linewidth]{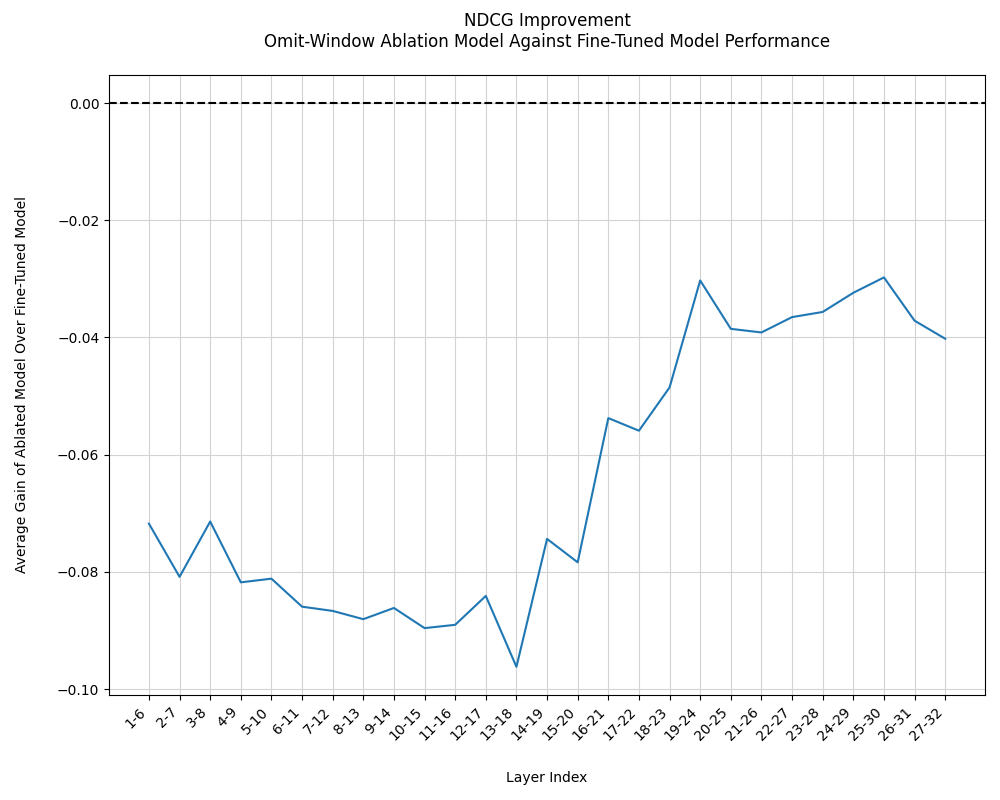}
    \caption{The performance difference between the fine-tuned model and ablated model, measured per-{\bf omitted-window} of size 6 in terms of NDCG. Window ablations show that earlier layers, roughly 1 through 20, are more performance-critical.}
    \label{fig:omit-window}
\end{figure}

\paragraph{Omit Ablations.}
Figures~\ref{fig:omit-head}--\ref{fig:omit-window} show the results of omit ablations across all three granularities, revealing which fine-tuned components are \textit{necessary} for strong ranking performance.
Omitting LoRA from the mid-network region (layers 10--18) consistently weakened performance the most, with layer 29 emerging as an isolated critical site.
Omitting the 13--18 window yielded the largest NDCG drop.

\begin{figure}
    \includegraphics[width=\linewidth]{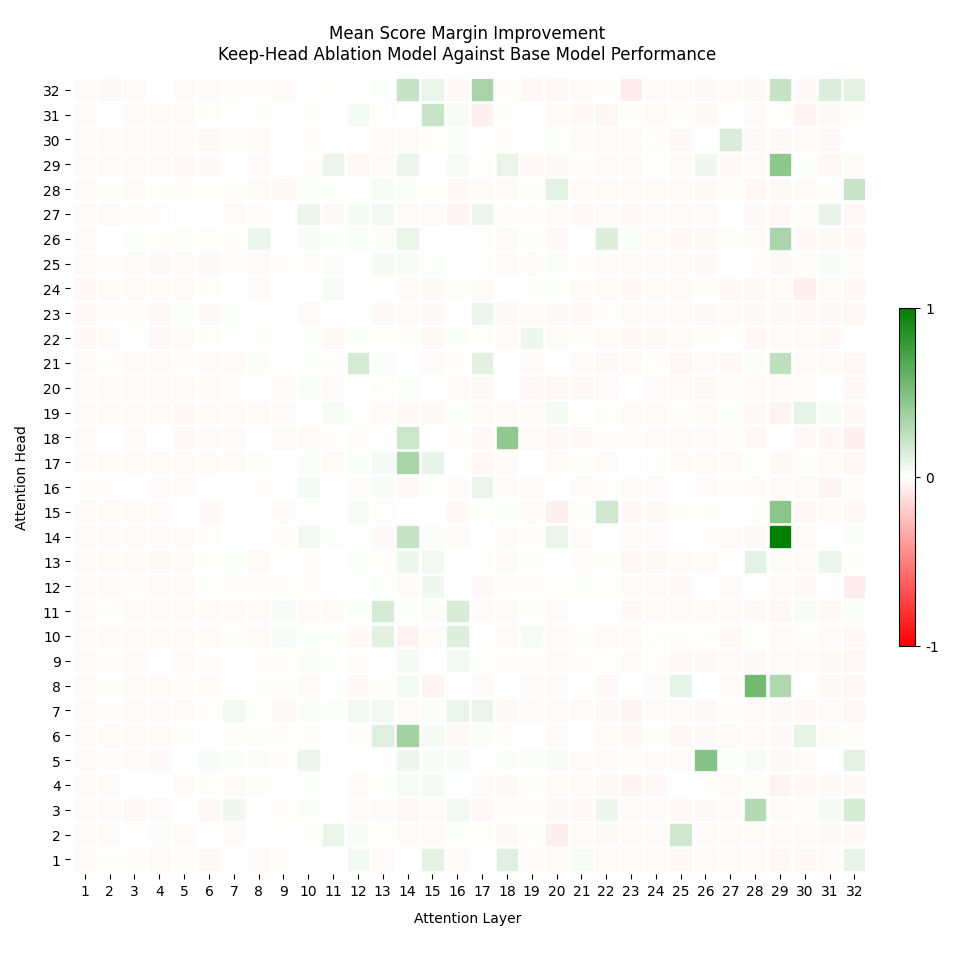}
    \caption{The performance difference between the base model and ablated model, measured per-{\bf kept-head} in terms of mean score margin. Scores are normalized by the maximum absolute value across all heads, such that the scale reflects relative importance in the network.
    Green indicates keeping the given head improves
performance and red indicates it hurts performance.
    Keeping fine-tuning on many of the heads in layers 10 through 18 and 25 through 32 benefits performance most.}
    \label{fig:keep-head}
\end{figure}

\begin{figure}[t]
    \includegraphics[width=\linewidth]{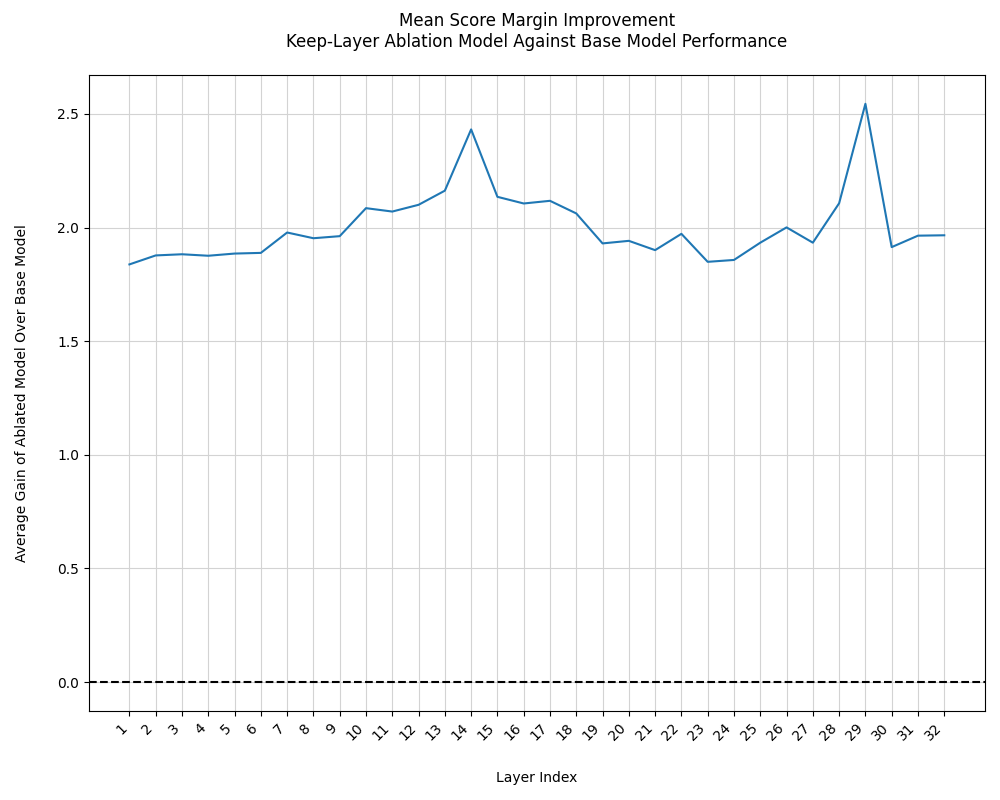}
    \caption{The performance difference between the base model and ablated model, measured per-{\bf kept-layer} in terms of mean score margin. Keeping fine-tuning produces roughly uniform performance gains throughout other than layers 14 and 29, which yielded notably higher improvement than others.}
    \label{fig:keep-layer}
\end{figure}

\begin{figure}[t]
    \includegraphics[width=\linewidth]{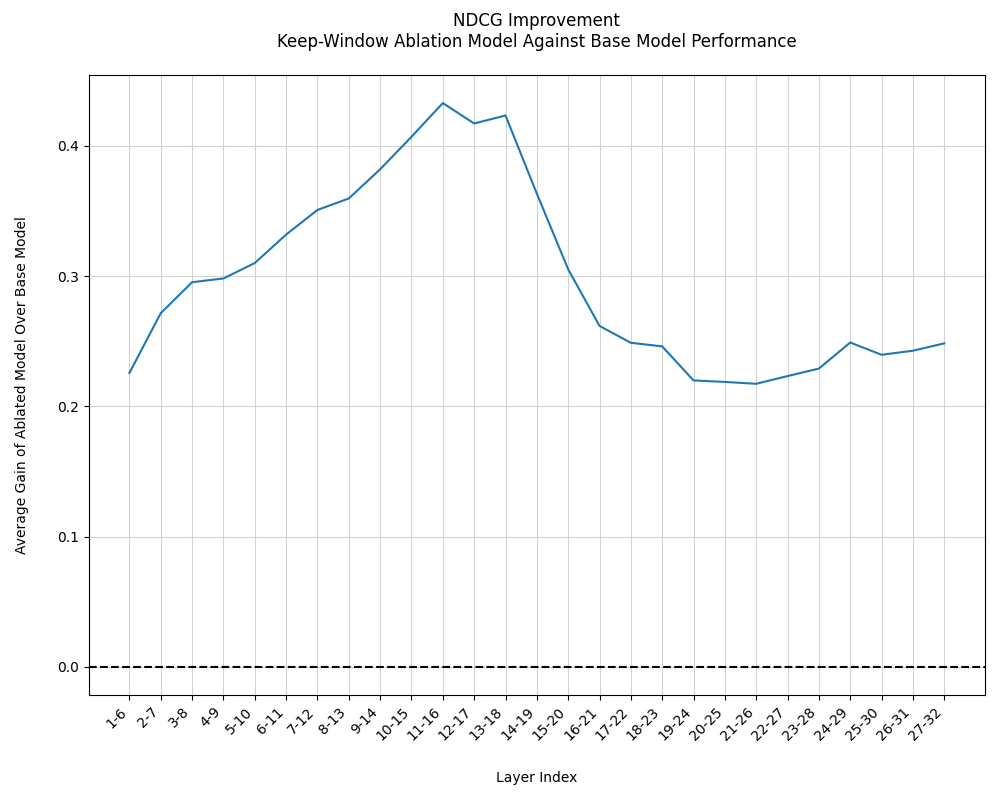}
    \caption{The performance difference between the base model and ablated model, measured per-{\bf kept-window} of size 6 in terms of NDCG. These ablations show that fine-tuning layers in the range 10 to 18 is generally most important for improving NDCG over the base model.}
    \label{fig:keep-window}
\end{figure}

\paragraph{Keep Ablations.}
Figures~\ref{fig:keep-head}--\ref{fig:keep-window} show the result of keep ablations across all three granularities, revealing which fine-tuned components are \textit{sufficient} for strong performance.
Heads in layers 10--18 and 25--32 contribute most when kept in isolation. Layer-level results show nearly uniform gains, with layers 14 and 29 standing out as high-impact sites.
Window ablations show the range 10--18 as most important, with windows in this region improving NDCG by over 0.4 points---more than half the difference between the base and fully LoRA fine-tuned model.

\paragraph{Combined Ablation Analysis.}
Both omit and keep ablations converge on a mid-network region, layers 10--18, as most critical, with layer 29 as a high-impact outlier. Regions that are sufficient but not necessary for performance gains suggest redundancy in the network: applying LoRA there in isolation helps, but omitting it from the otherwise fully LoRA fine-tuned model has little effect.

\section{Feature Attention Experiments}

The second experimental question is whether axiomatic attention patterns are learned during LoRA fine-tuning (RQ2).
To study this, we measure how much attention mass each head allocates to interpretable token-pair features before and after LoRA fine-tuning.
Analyzing this across all heads, layers, and windows reveals which attention behaviors emerge through LoRA and where in the network.

\subsection{Methodology}

\paragraph{Axiomatic Attention Features.}
Following the axiomatic IR framework of \citet{fang2005axiomatic}, we define several interpretable token-pair features associated with retrieval relevance.

\begin{itemize}
    \item \textbf{Lexical Matching}: A token pair $(i, j)$ is considered a lexical match if the words containing tokens $i$ and $j$ are identical under lowercased, whitespace-trimmed string comparison.


    \item \textbf{Rarity Sensitivity}: A token is considered \textit{rare} based on the inverse document frequency (IDF \citep{sparckjones1972}) of the word it belongs to, computed over a sample of 500k documents from the MS MARCO training set. A word is labeled rare if its IDF is above an empirical threshold (see \S\ref{app:idf}).

    \item \textbf{Document-Query Interaction}: A token pair exhibits document-query interaction when one token belongs to the document segment and the other belongs to the query segment. Since RankLLaMA is decoder-only and the input ordering places the query before the document, interaction is constrained to document tokens attending query tokens.
\end{itemize}

\paragraph{Compositional Features.}
In addition to the three individual features listed above, we also explore \textit{compositional} token-pair features that combine multiple conditions simultaneously. A pair $(i, j)$ satisfies a compositional feature only if it meets all criteria at once---for example, a rare document token attending to a lexically matching query token. Such compositions more closely mirror the interdependent signals that underlie classical retrieval models: BM25 jointly rewards lexical matching, document-query co-occurrence, and IDF weighting \citep{robertson2009bm25}, and axiomatic IR theory similarly recognizes such relevance signals interact rather than operate independently \citep{fang2005axiomatic}. Compositional features therefore allow us to ask not just whether individual relevance signals appear in attention, but whether fine-tuning attends richer interdependent features consistent with established IR axioms.

\paragraph{Normalized Feature Attention.}
In practice, a small number of tokens absorb a disproportionate share of per-head attention, regardless of content --- a phenomenon known as the \textit{attention sink} \citep{xiao2023efficient}. In our setting, we observe empirically that the sink is consistently and exclusively the first token in the sequence. This is problematic for measuring how much attention falls on a given feature, as genuine feature-driven attention is confounded by mass trivially absorbed by the sink.

We define \textit{Normalized Feature Attention}, a new metric for detecting attention patterns to address this. Let $P_f$ denote the set of all token pairs $(i, j)$ satisfying feature $f$. We exclude any pair where the attended token $j$ is the sink token (i.e., $j = 0$). The normalized feature attention of head $h$ to feature $f$ on a single input is then defined as:

\begin{equation}
\label{eq:nfa}
N^h_f =
\frac{
\displaystyle\sum_{(i,j)\in P_f} a^h_{ij}
}{
\displaystyle\sum_i \sum_{j>0} a^h_{ij}
}
\end{equation}

\noindent where $a^h_{ij}$ is the attention weight from token $i$ to token $j$ in head $h$. The final reported value is the mean of $N^h_f$ across all sequences in the evaluation set. By normalizing over non-sink attention mass, the metric measures the proportion of meaningful attention allocated to the feature of interest. We provide visualizations of sink tokens and normalized feature attention in \S\ref{app:normalized-attn}.

\paragraph{Evaluation Metrics.}
We compute average normalized feature attention, per-head, over 1{,}000 random relevant query-document pairs from MS MARCO \citep{bajaj2016msmarco}.

To isolate the effects of fine-tuning, we compare the fully LoRA fine-tuned model against the base model. For each head and feature, we report the signed difference in normalized feature attention:

\begin{equation}
\Delta N^h_f = N^h_{f,\mathrm{FT}} - N^h_{f,\mathrm{base}}
\end{equation}

\noindent where positive values indicate fine-tuning increased attention to the feature and negative values indicate decreased attention relative to the base model.

\subsection{Results}

\begin{figure}[h!]
    \centering
    \includegraphics[width=0.95\linewidth]{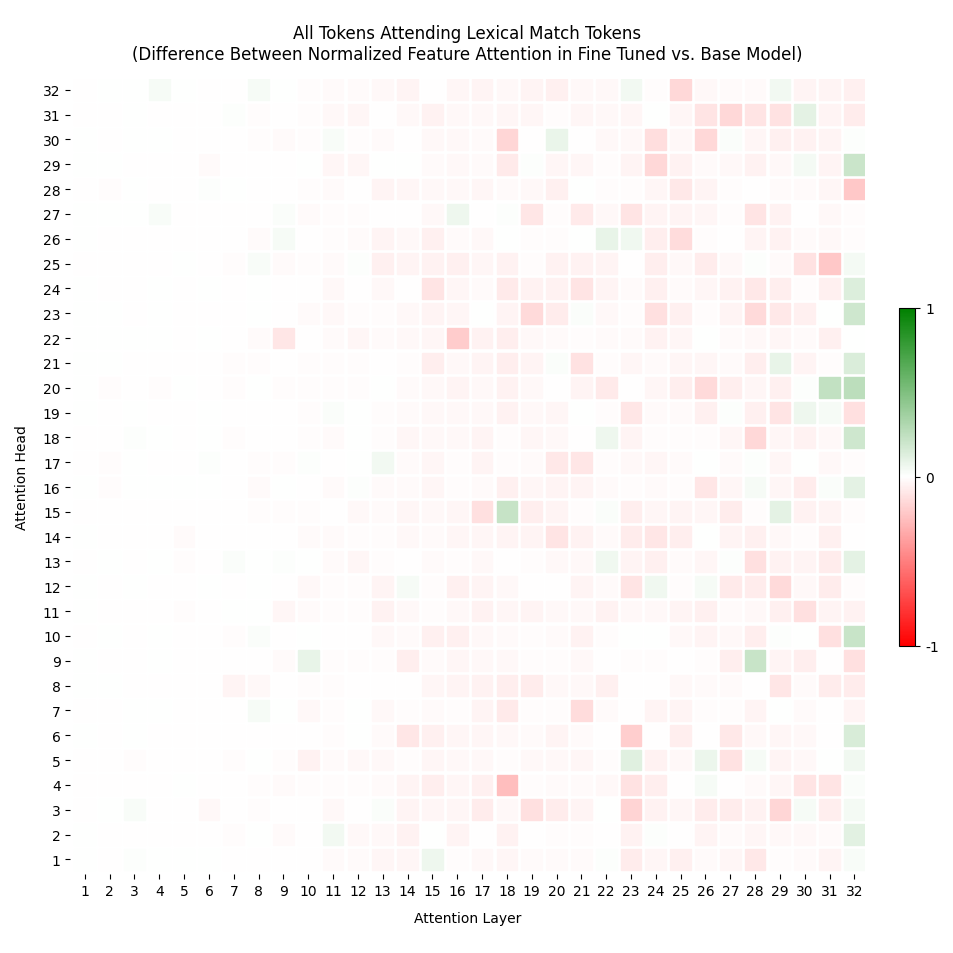}
    \caption{The difference between normalized lexical matching attention, measured per-{\bf head}, before and after fine-tuning. Attention to lexical matching generally decreases in the fine-tuned model, especially for layers 11 on. Lexical matching notably increases in the last layer relative to the base model.}
    \label{fig:lexical-matching}
\end{figure}

\begin{figure}[h!]
    \centering
    \includegraphics[width=0.95\linewidth]{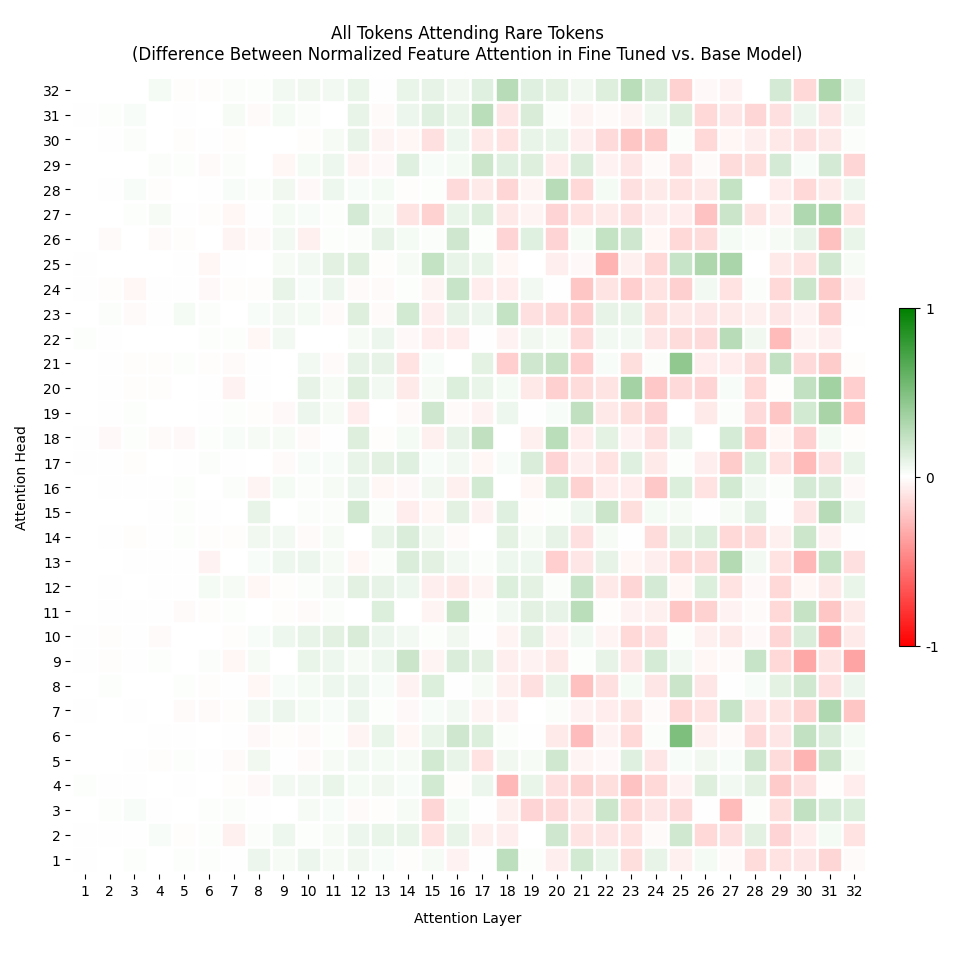}
    \caption{The difference between normalized rarity-sensitive attention, measured per-{\bf head}, before and after fine-tuning. Rarity-sensitive attention generally increased in the fine-tuned model for layers 8 through 19 and decreased for layers 20 through 32. Despite general trends at the window level, rarity sensitivity fluctuates dramatically per-head.}
    \label{fig:rarity}
\end{figure}

\begin{figure}[h!]
    \centering
    \includegraphics[width=0.95\linewidth]{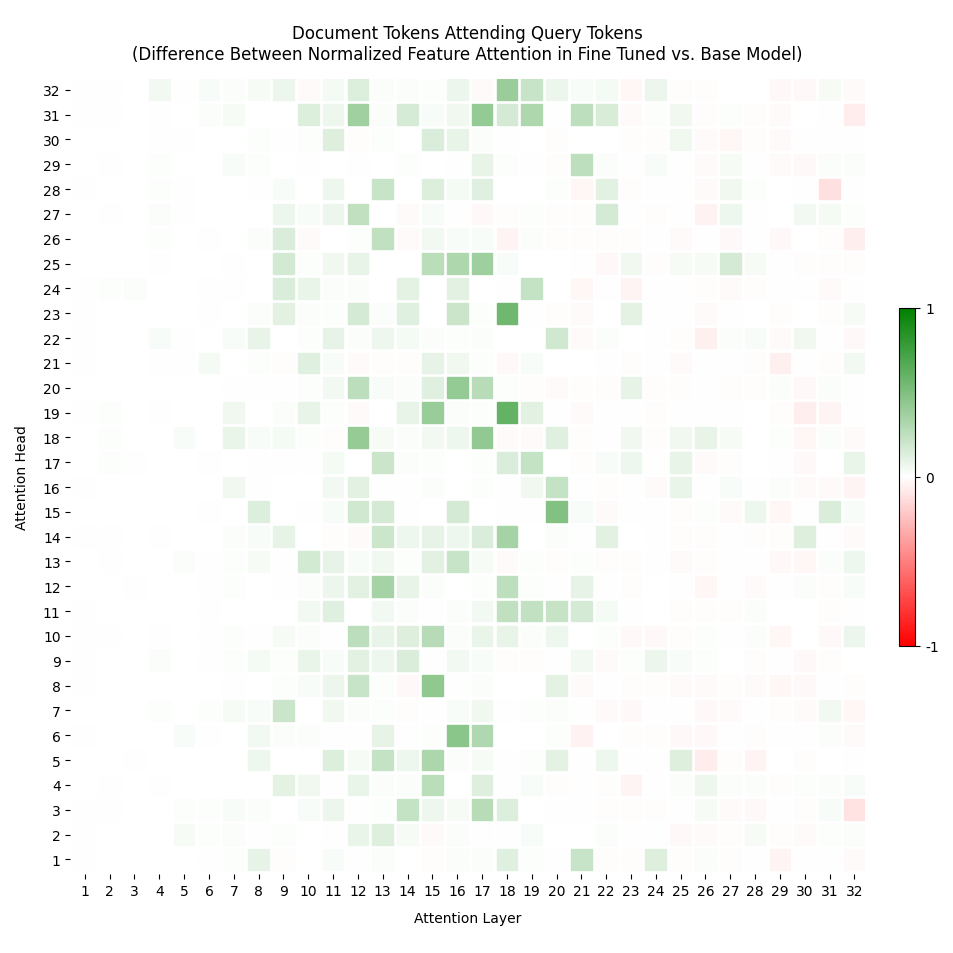}
    \caption{The difference between normalized document-query attention, measured per-{\bf head}, before and after fine-tuning. The fine-tuned model exhibits document-query attention more than the base model in layers 8--22.}
    \label{fig:document-query}
\end{figure}

Figures~\ref{fig:lexical-matching}--\ref{fig:document-query} show the change in normalized feature attention between the base and fine-tuned models across all heads for each of the three axiomatic features.
Green indicates the given head attends to the feature more after fine-tuning and red indicates it attends less.
Attention to lexical matching generally decreases after fine-tuning, particularly from layer 11 onward, with the exception of the final layer. We believe the increase in the final layer is an artifact of self-attention collapsing toward the identity, as we observe many tokens attend predominantly to themselves rather than genuine cross-sequence term matches. Rarity sensitivity follows a directional shift: attention to rare terms increases in layers 8--19 and decreases from layer 20 onward, although per-head variation within layers is high despite consistent layer-wise trends. Document-query interaction shows the clearest learned signal, with the fine-tuned model attending across segment boundaries more than the base model throughout layers 8--22.
Normalized feature attention heatmaps for the base and fine-tuned models individually are available in \S\ref{app:heatmaps}.

Compositional features combining these signals also exhibit interpretable trends in learned attention, which we discuss in \S\ref{app:compositions}.

\section{Discussion}

We now use the results of the ablation and attention experiments to explore the relationship between LoRA-driven performance gains and interpretable trends in learned attention (RQ3).

Our ablations show that given LoRA fine-tuned MLPs throughout the network, applying LoRA attention updates to layers 10–18 is \textit{necessary} to match the performance of full LoRA application and \textit{sufficient} to recover more than half of its performance gain over the base model.
Our attention analysis shows fine-tuning decreases lexical matching after layer 11, increases rarity sensitivity in layers 8–19, decreases it in 20–32, and increases document-query interaction in layers 8–22.

For all three axiomatic features we investigated, especially rarity sensitivity and document-query interaction, the regions where LoRA fine-tuning most changes normalized feature attention align with the regions where LoRA most improves performance.
Table~\ref{tab:spearman} quantifies these relationships, demonstrating strong correlations for both rarity sensitivity and document-query interaction. The weaker correlation for lexical matching likely reflects two limitations of how we examine matching: it cannot capture semantic matches between non-identical tokens \citep{fang2006semantic}, and many heads score artificially high for lexical matching due to token-wise self-attention. Empirical review of attention heatmaps suggests dedicated semantic matching heads are learned, which we leave for future study.

\begin{table}[H]
\centering
\small
\begin{tabular}{lcc}
\toprule
\textbf{Feature} & \textbf{Keep} $\rho$ & \textbf{Omit} $\rho$ \\
\midrule
Lexical Matching   & 0.37  & $-$0.63 \\
Rarity Sensitivity            & 0.92  & $-$0.89 \\
Document-Query Interaction      & 0.71  & $-$0.68 \\
\bottomrule
\end{tabular}
\caption{Spearman correlation between the effect of per-window ablations on NDCG performance and per-window change in normalized feature attention after fine-tuning, computed over sliding windows of size 6. Keep correlations are positive because windows that learn stronger feature attention also produce larger performance gains when kept in isolation; omit correlations are negative because those same windows produce larger performance drops when removed. For correlations over other sliding window sizes, see \S\ref{app:window-sizes}.}
\label{tab:spearman}
\end{table}

We also find certain compositional features have matching or higher correlation with performance gains than individual features, suggesting interdependency between axiomatic properties is critical (see \S\ref{app:compositions}).
This result agrees with top classical ranking strategies such as BM25 \citep{robertson2009bm25} that leverage all three features, and supports prior work discovering interdependent signals in neural rerankers \citep{tanya2025probing, lu2025pathway}.

\section{Conclusion}

This paper studies where LoRA fine-tuning learns relevance in RankLLaMA, a decoder-based reranker, and whether regions where applying LoRA is performance-critical coincide with learned axiomatic attention patterns. To support this analysis, we introduce normalized feature attention, a metric that controls for attention sink mass and enables reliable measurement of attention patterns. Across head-, layer-, and window-level ablations using a keep/omit design that separately identifies sufficient and necessary components, we find gains in ranking performance are primarily concentrated in a compact set of mid-network layers. These critical regions are strongly correlated with LoRA-driven attention to interpretable features such as rarity sensitivity and document-query interaction. Compositional features combining these signals are even more predictive of performance gains, mirroring the joint relevance signals of classical models such as BM25. These results support an axiomatically grounded correlational account of where and how relevance-oriented behavior is learned during LoRA fine-tuning, suggesting the opportunity for targeted fine-tuning in neural reranking models.

\section{Limitations}

We only analyze matching through direct lexical comparison, which leaves the question of semantic matching unresolved.
Our experiments are conducted on a single model (RankLLaMA-7B) and evaluated on MS MARCO with small candidate sets and random negatives, which limits generalizability to other architectures and retrieval settings. We partially address generalization in \S\ref{app:generalization}, where we replicate our ablation and attention findings on a second reranking objective. While the correlation between learned attention patterns and performance gains supports an axiomatic mechanistic interpretation, attention trends alone do not constitute a causal relevance mechanism. Future work incorporating multiple models, more features, harder evaluation negatives, and causal intervention methods is necessary for a complete causal account of how relevance is learned during LoRA.

\section{Ethical Considerations}

This work presents interpretability analysis of a publicly available neural reranking model (RankLLaMA-7B) trained on MS MARCO, a widely used dataset across NLP. All experiments use publicly available models and data; no new data collection, human subjects, or sensitive information is involved. Our findings, which interpret attention patterns in regions where LoRA is critical for performance, may inform more efficient fine-tuning practices in the future, which we consider a net positive for engineers and the environment. We see no foreseeable harm arising from this research.

\section*{Acknowledgments}
This work was supported in part by the Center for Intelligent Information Retrieval. Any opinions, findings and conclusions or recommendations expressed in this material are those of the authors and do not necessarily reflect those of the sponsor.

\clearpage

\bibliographystyle{plainnat}
\bibliography{main_refs}

\clearpage

\appendix

\section{Generalizing to a Second Model}
\label{app:generalization}

To assess whether our findings generalize beyond a single reranking objective, we repeat our keep/omit ablation and attention experiments on the RankLLaMA-7B \textit{document} variant, a separately trained reranker optimized for document retrieval rather than passage retrieval.

We observe the same qualitative behavior as the \textit{passage} variant. Given LoRA fine-tuned MLPs throughout the network, critical LoRA attention updates are concentrated in a compact mid-network region (approximately layers 7--16). Keeping only this region recovers over 70\% of the fully LoRA fine-tuned model's NDCG gain over the base model, and omitting it degrades performance more than elsewhere in the network. Document-query interaction again aligns strongly with this performance-critical region. While the prevalence of individual axiomatic features varies between the two reranking objectives, our central finding is consistent: given LoRA fine-tuned MLPs throughout the network, LoRA attention updates concentrate relevance-critical behavior in a compact mid-network region associated with interpretable and axiomatic attention patterns.

\section{Baseline Model Performance}
\label{app:control}

When evaluating different ablations, we compute performance relative to the base model for keep ablations and relative to the fine-tuned model for omit ablations. For layer-wise and window-wise ablations, the evaluation set is 50 queries $ \times $ 100 candidate documents, resulting in the baseline performance metrics listed in Table~\ref{tab:performance}.
To reduce the computational cost of evaluating 32 $\times$ 32 model variants, head-wise ablations are evaluated on 10 candidate documents per query, yielding the baseline performance values shown in Table~\ref{tab:performance-small-set}.

\begin{table}[h!]
\centering
\small
\begin{tabular}{lcc}
\toprule
 & \textbf{NDCG} & \textbf{Mean Score Margin} \\
\midrule
\textbf{Base Model} & 0.437 & -0.106 \\
\textbf{Fine-Tuned Model} & 0.970 & 8.866 \\
\midrule
\textbf{Performance Gain} & 0.533 & 8.972 \\
\bottomrule
\end{tabular}
\caption{Performance of the base model and fine-tuned model evaluated over 50 queries $\times$ 10 documents.}
\label{tab:performance-small-set}
\end{table}

\vfill

\section{IDF Threshold for Rare Tokens}
\label{app:idf}

We define a word as rare if its IDF, calculated over a sample of 500k documents from the MS MARCO training set, exceeds an empirical threshold. We set this threshold at approximately the IDF score of the 180th most frequent word in the corpus, consistent with the size of many common stopword lists. The vast majority of unique words in a large corpus are informative, and this threshold ensures stopwords are not marked as rare while ensuring actually informative words are considered.

\section{Normalized Feature Attention}
\label{app:normalized-attn}

\begin{figure*}
    \centering
    \includegraphics[width=0.42\textwidth]{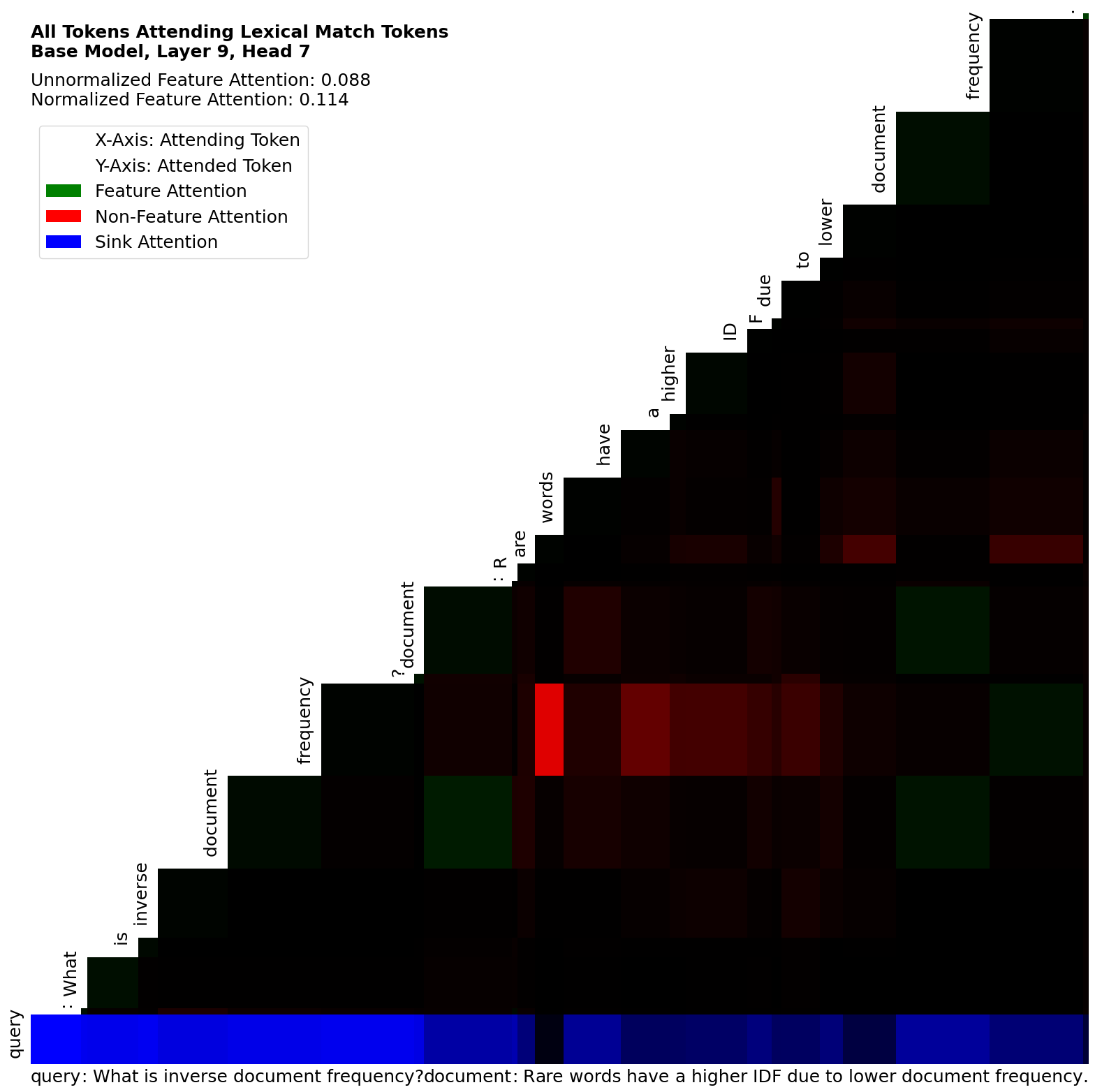}
    \hfill
    \includegraphics[width=0.42\textwidth]{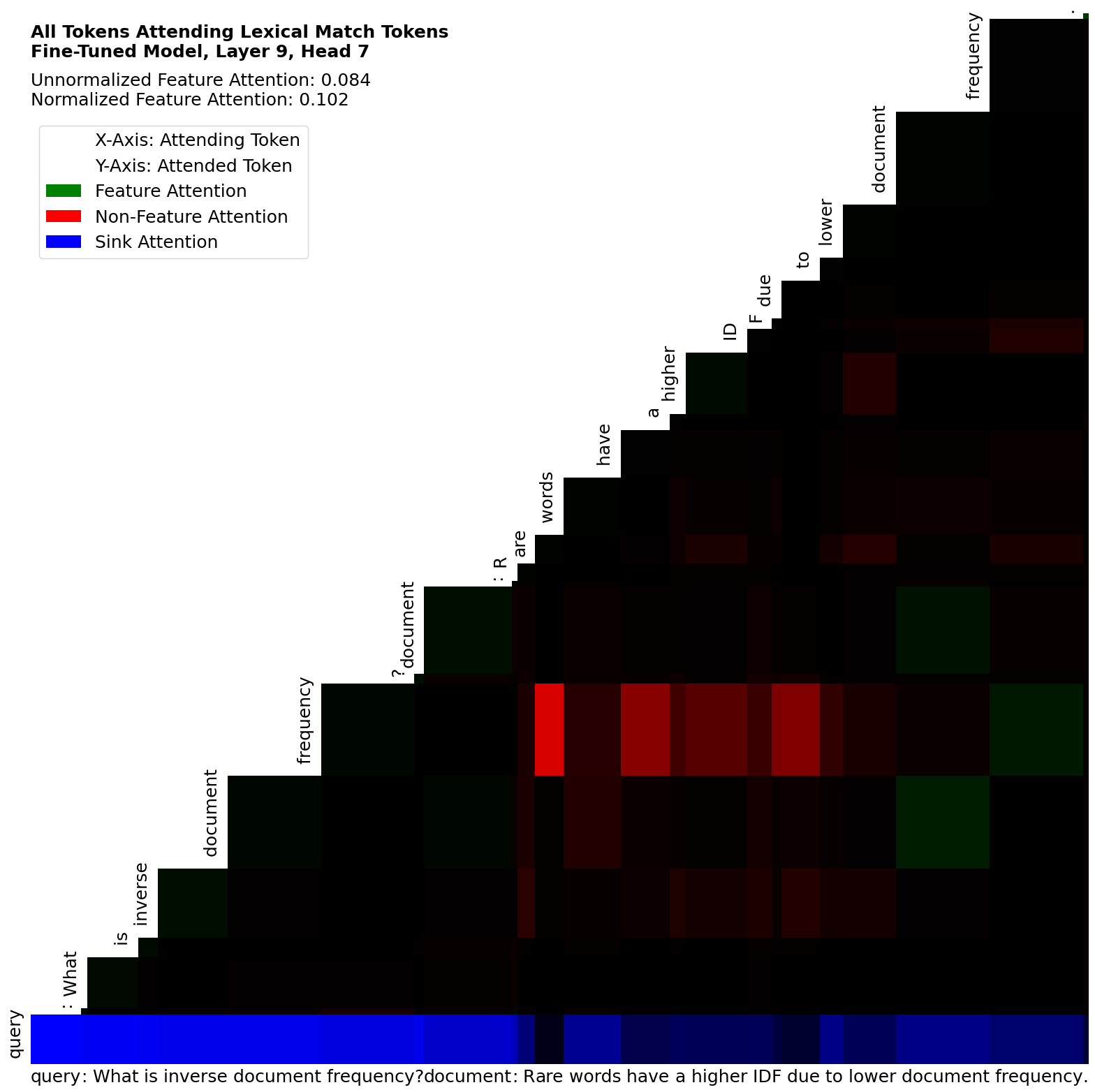}
    \caption{Lexical match attention in the base model (Left) vs. fine-tuned model (Right).}
    \label{fig:lexical-heatmaps}
\end{figure*}

\begin{figure*}
    \centering
    \includegraphics[width=0.42\textwidth]{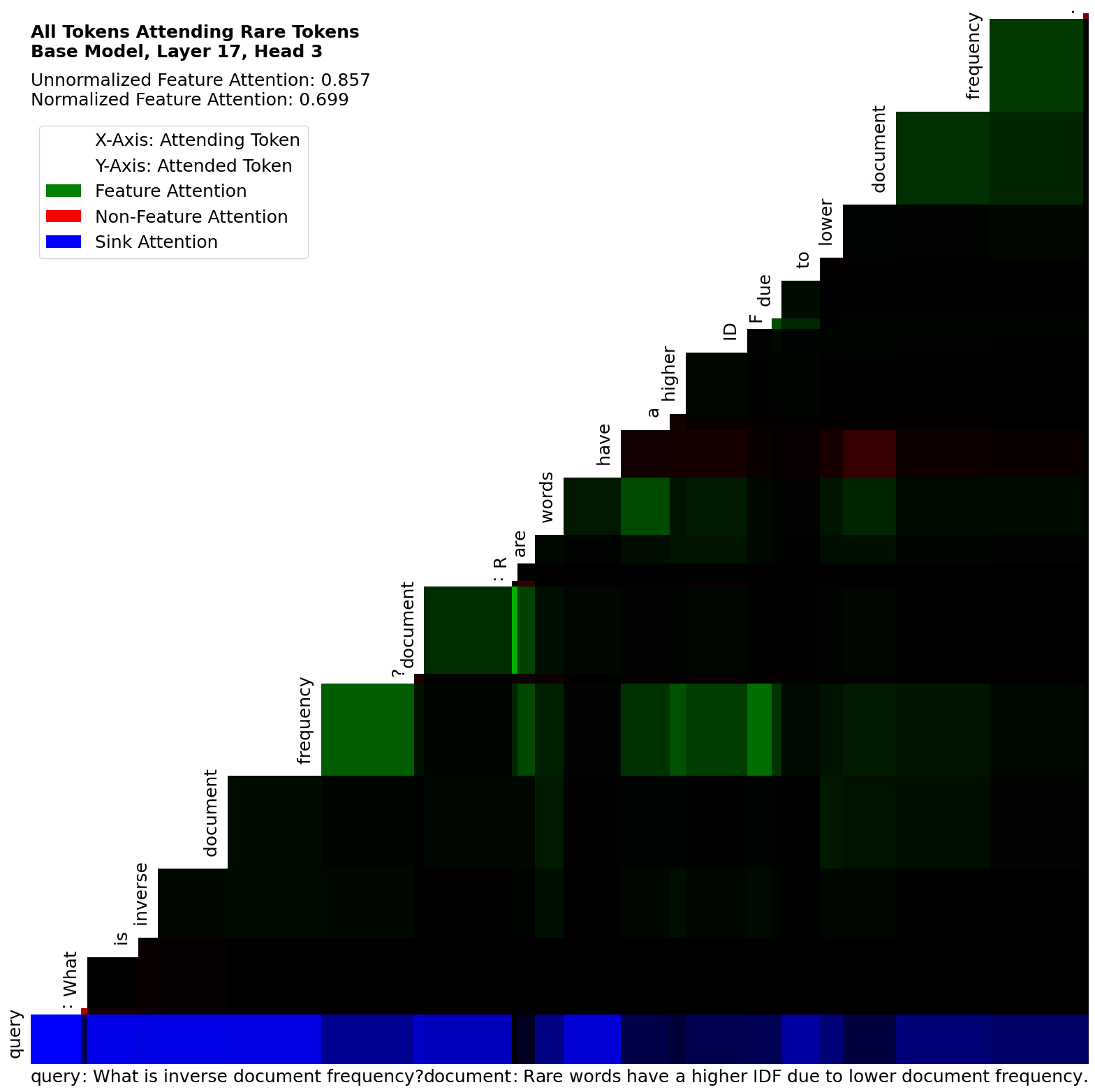}
    \hfill
    \includegraphics[width=0.42\textwidth]{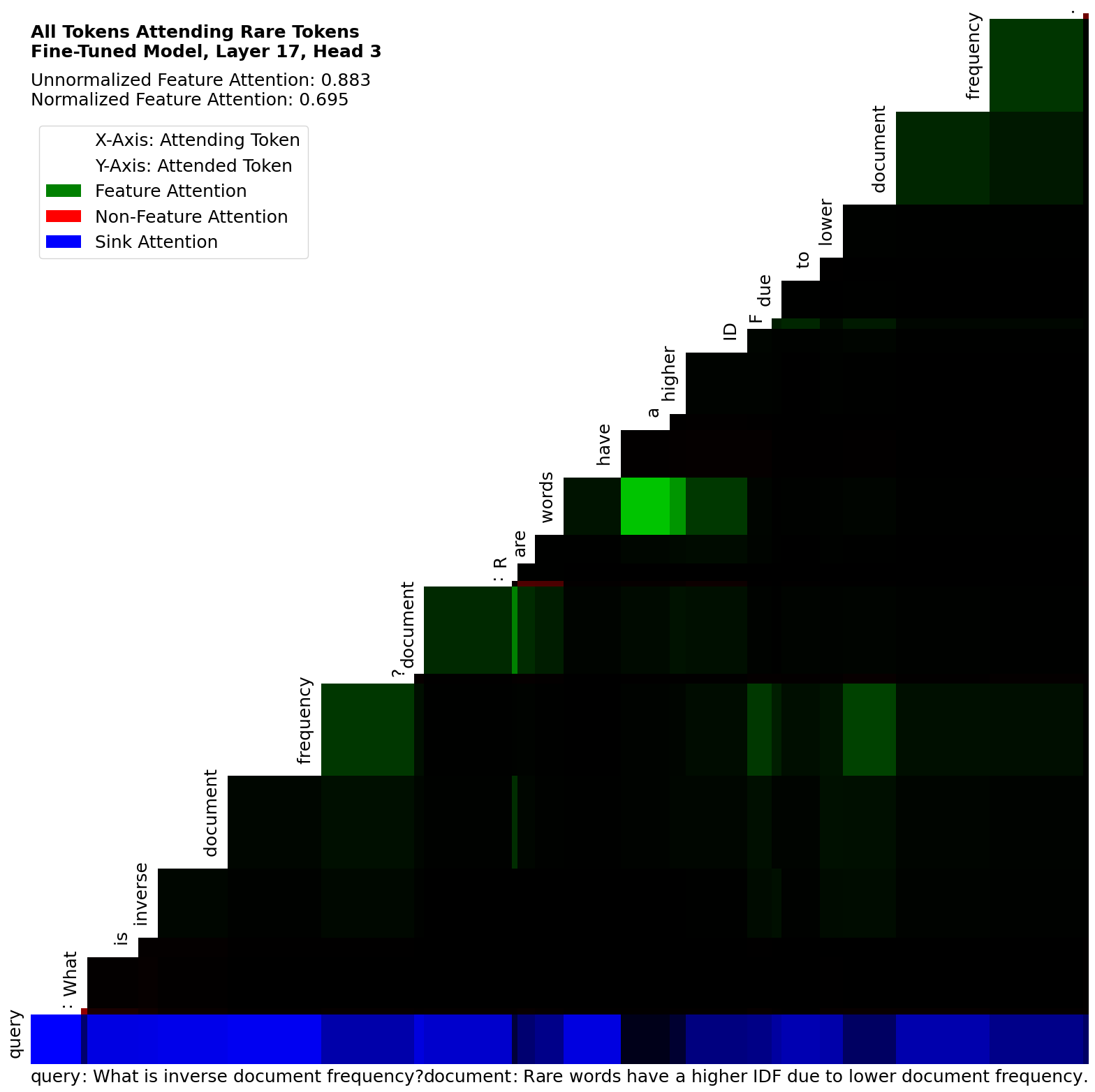}
    \caption{Rarity-sensitive attention in the base model (Left) vs. fine-tuned model (Right).}
    \label{fig:rare-heatmaps}
\end{figure*}

\begin{figure*}
    \centering
    \includegraphics[width=0.42\textwidth]{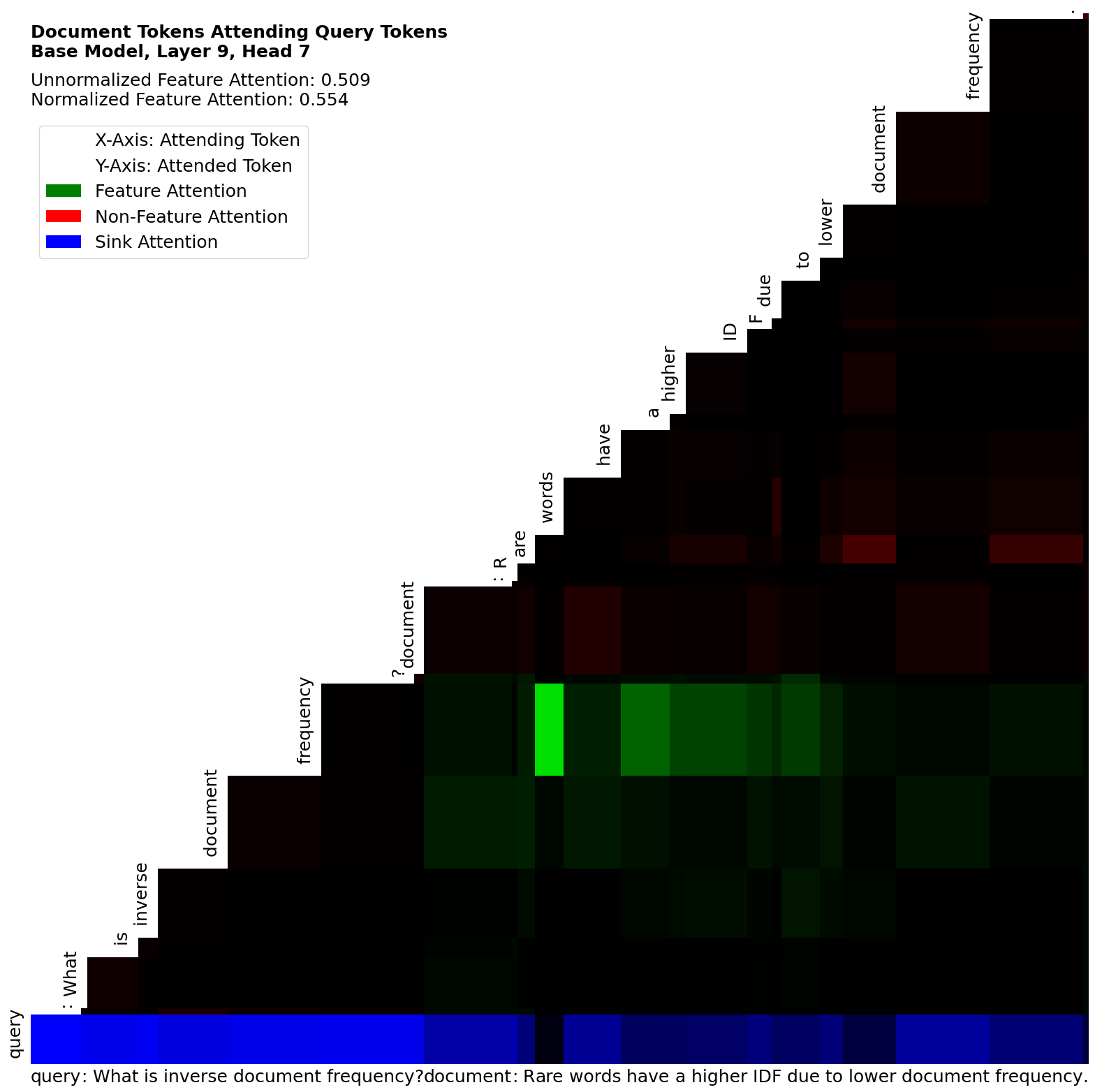}
    \hfill
    \includegraphics[width=0.42\textwidth]{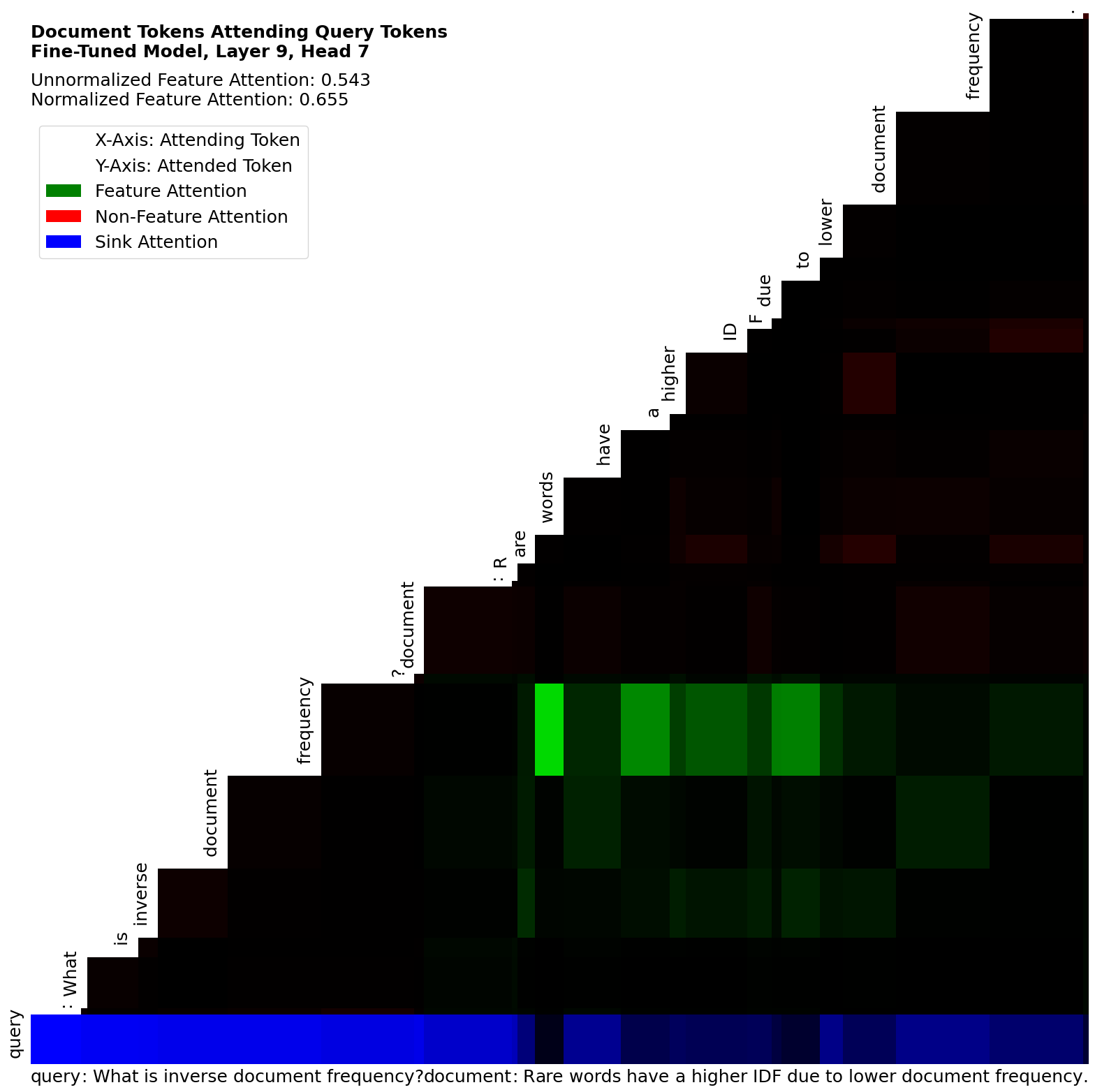}
    \caption{Document-query attention in the base model (Left) vs. fine-tuned model (Right).}
    \label{fig:doc-query-heatmaps}
\end{figure*}

To build an intuition for axiomatic features and understand the role of sink-mass normalization, we provide feature attention visualizations for an example document-query pair in Figures~\ref{fig:lexical-heatmaps}-\ref{fig:doc-query-heatmaps}.

\section{Additional Attention Heatmaps}
\label{app:heatmaps}

\begin{figure}
    \centering
    \includegraphics[width=0.8\linewidth]{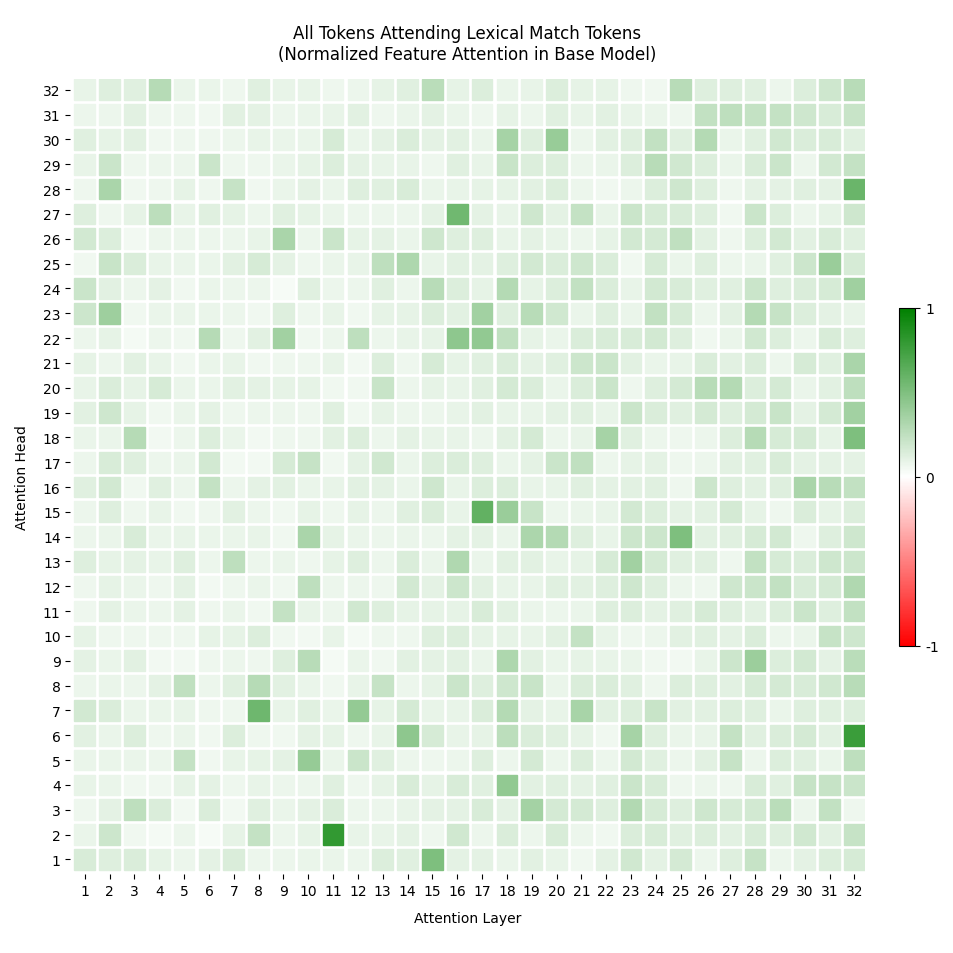}
    \caption{Normalized attention to lexical matching, measured per-head, in the Base Model.}
    \label{fig:lexical-matching-base}
\end{figure}

\begin{figure}
    \centering
    \includegraphics[width=0.8\linewidth]{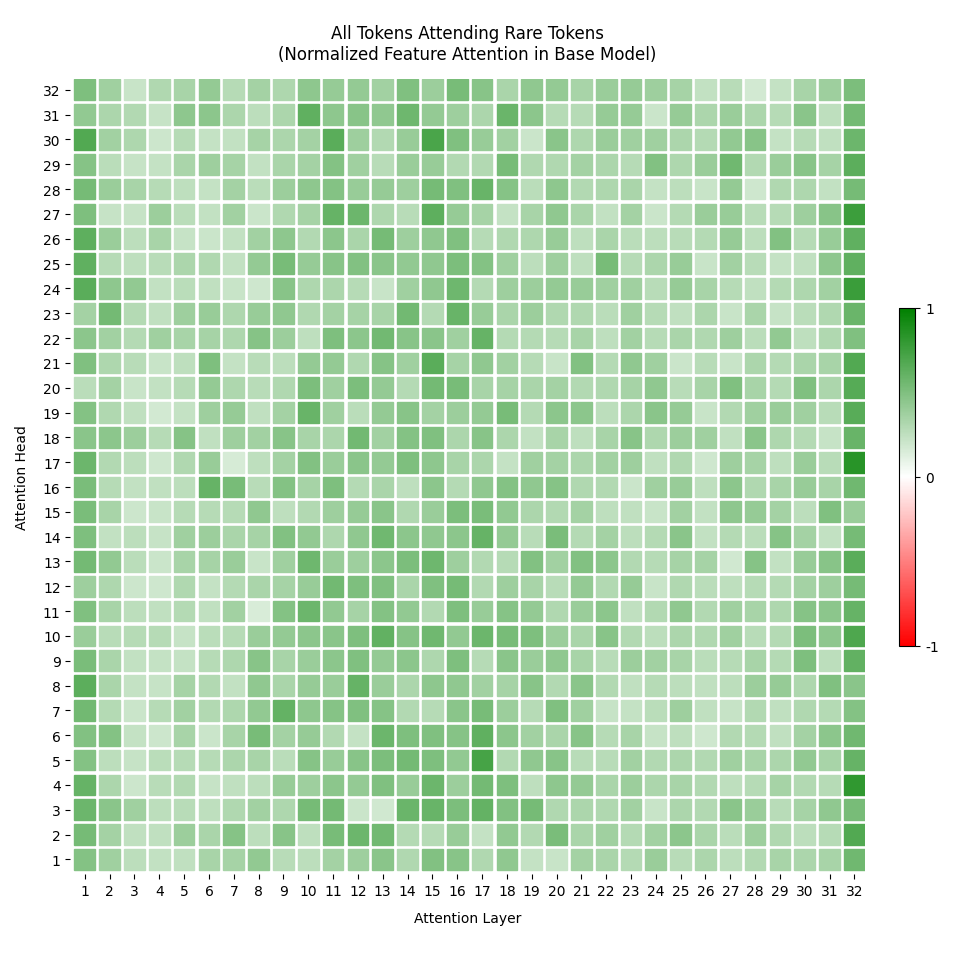}
    \caption{Normalized attention to rare terms, measured per-head, in the Base Model.}
    \label{fig:rarity-matching-base}
\end{figure}

\begin{figure}
    \centering
    \includegraphics[width=0.8\linewidth]{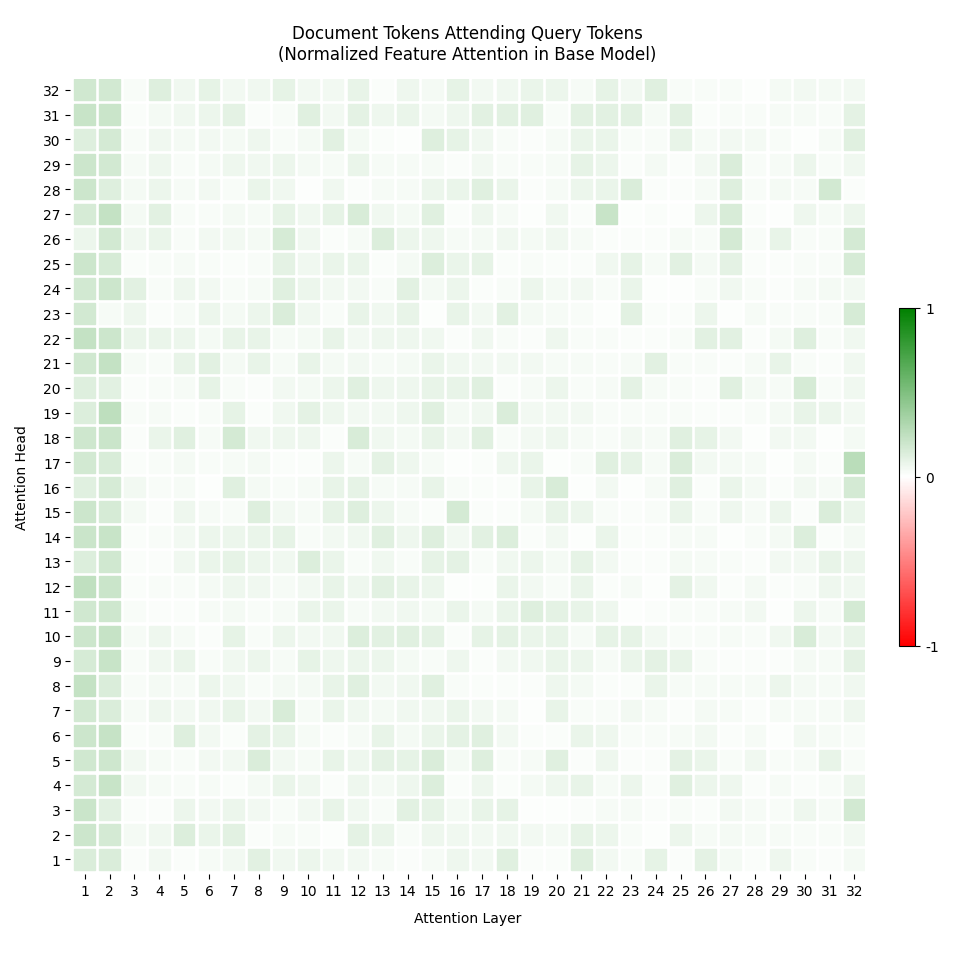}
    \caption{Normalized attention across document-query segments, measured per-head, in the Base Model.}
    \label{fig:doc-query-base}
\end{figure}

\begin{figure}
    \centering
    \includegraphics[width=0.8\linewidth]{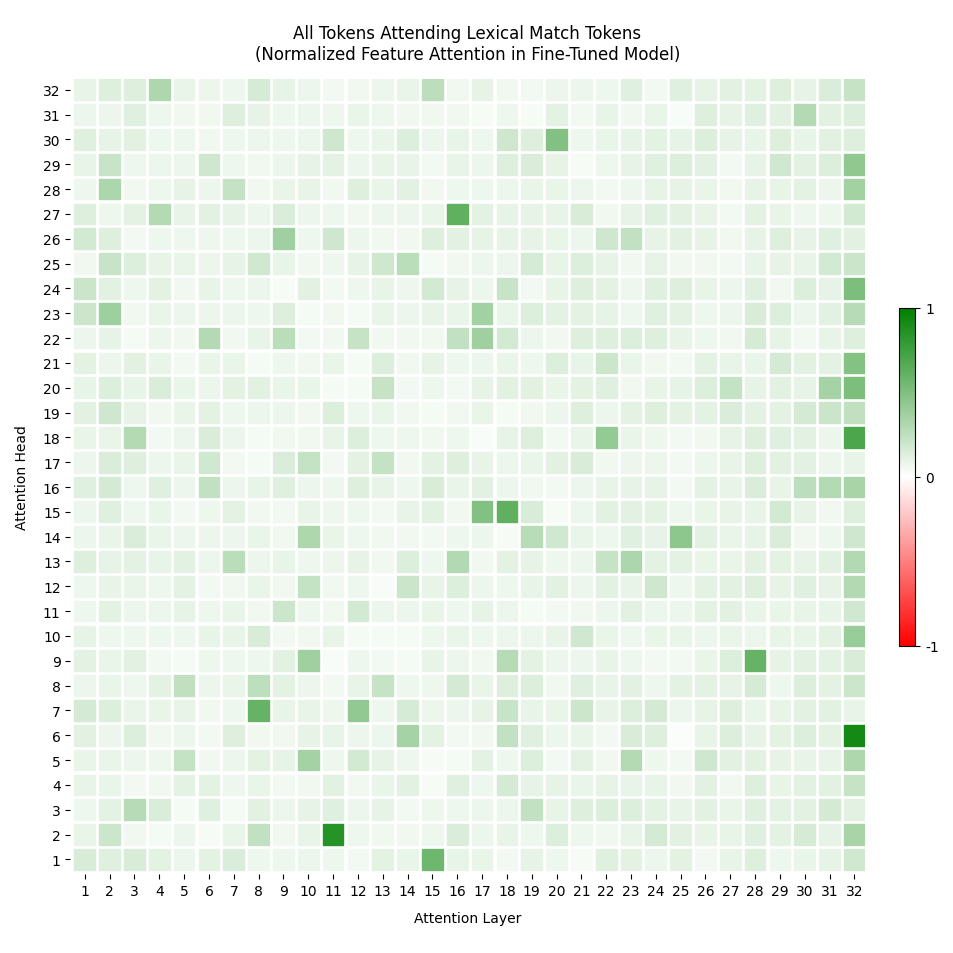}
    \caption{Normalized attention to lexical matching, measured per-head, in the fine-tuned model.}
    \label{fig:lexical-matching-ft}
\end{figure}

\begin{figure}
    \centering
    \includegraphics[width=0.8\linewidth]{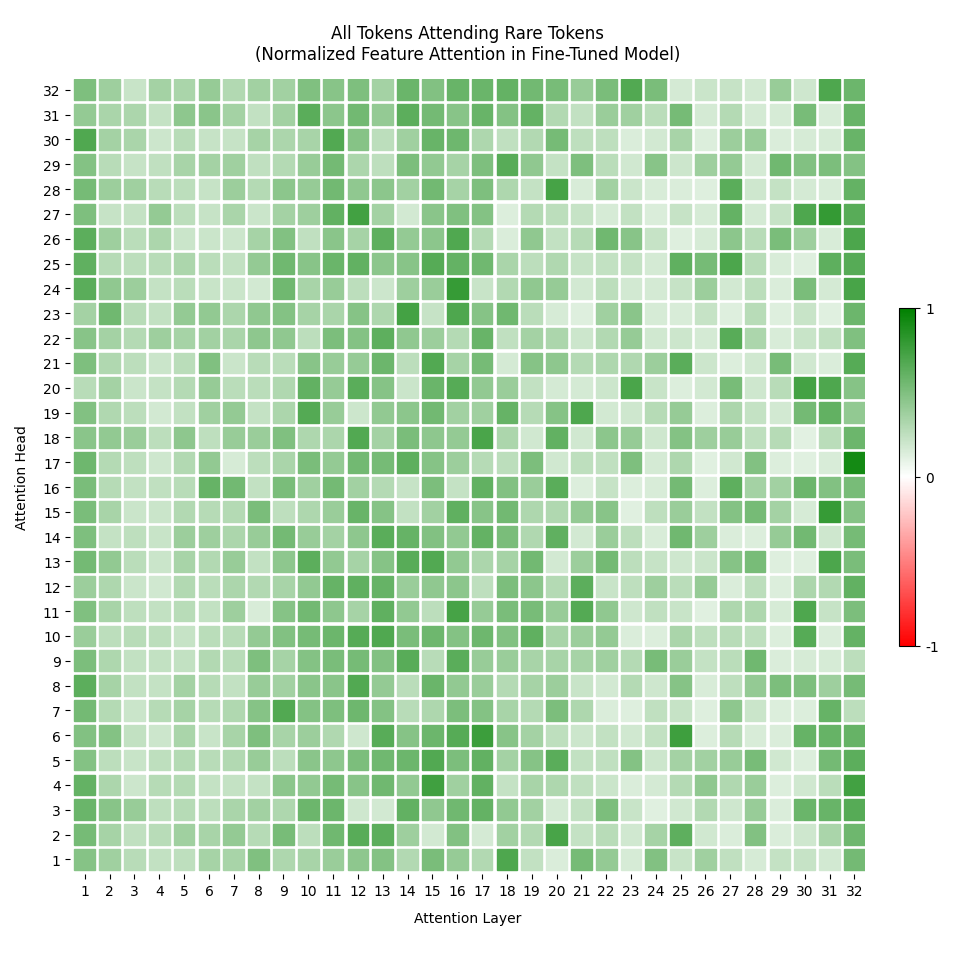}
    \caption{Normalized attention to rare terms, measured per-head, in the fine-tuned model.}
    \label{fig:rarity-matching-ft}
\end{figure}

\begin{figure}
    \centering
    \includegraphics[width=0.8\linewidth]{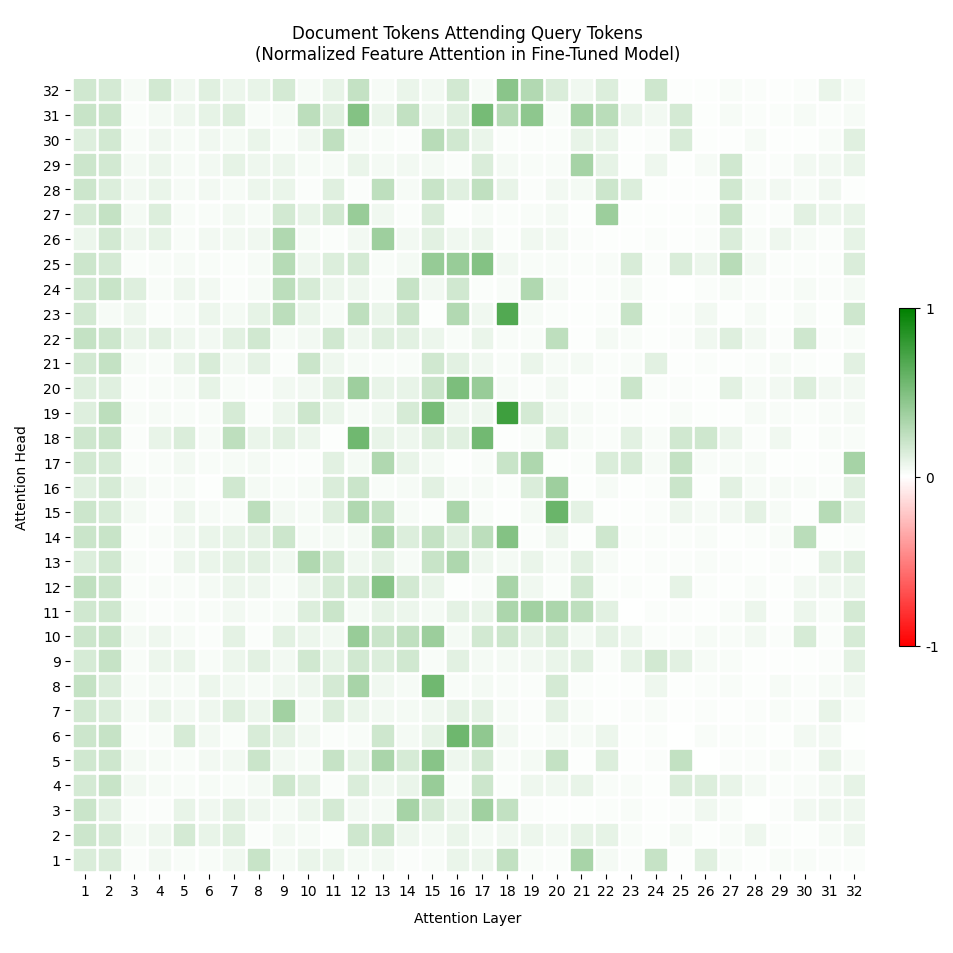}
    \caption{Normalized attention across document-query segments, measured per-head, in the fine-tuned model.}
    \label{fig:doc-query-ft}
\end{figure}

Figures~\ref{fig:lexical-matching-base}-\ref{fig:doc-query-ft} show normalized feature attention, per-head, for the base and fine-tuned models, complementing the delta heatmaps in Figures~\ref{fig:lexical-matching}-\ref{fig:document-query}.

\section{Compositional Feature Analysis}
\label{app:compositions}

In addition to the three core axiomatic features we investigate in the paper, there are many ways to combine these into more selective features, for example, query tokens attending rare tokens. Most of these compositional features have distinct trends in where they are learned during fine-tuning.

\begin{figure*}
    \centering
    \includegraphics[width=0.8\linewidth]{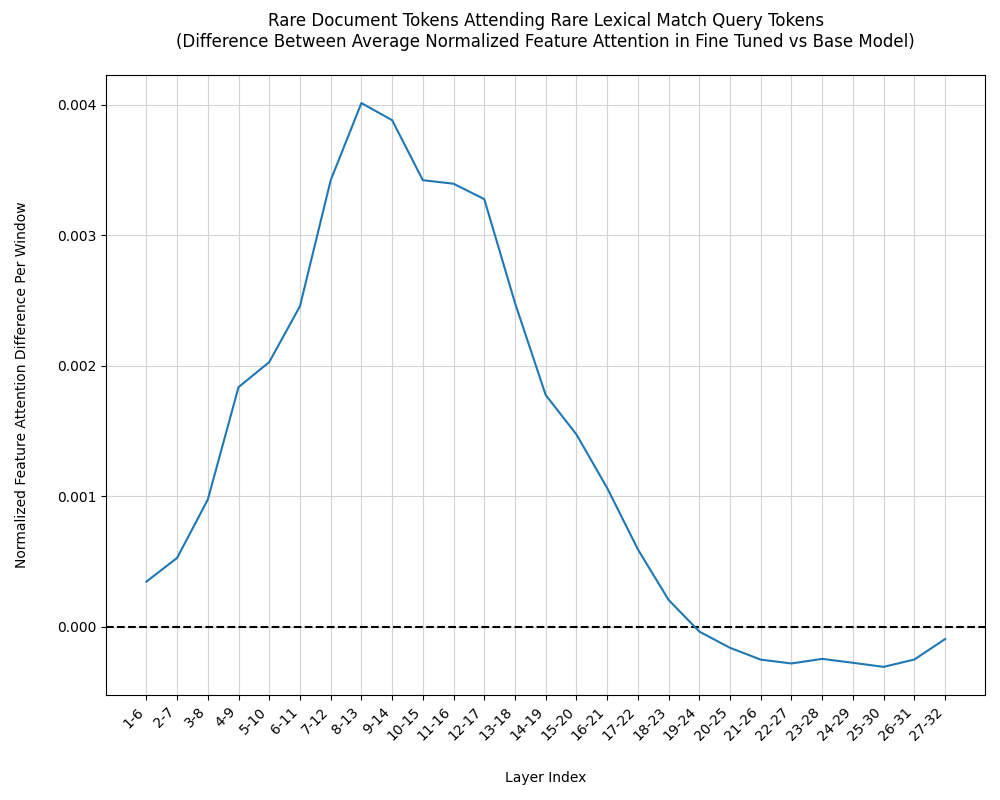}
    \caption{Change in normalized feature attention for rare document tokens attending rare lexical match query tokens, averaged over sliding windows of size 6, between the base and fine-tuned models.}
    \label{fig:composition1}
\end{figure*}

\begin{figure*}
    \centering
    \includegraphics[width=0.8\linewidth]{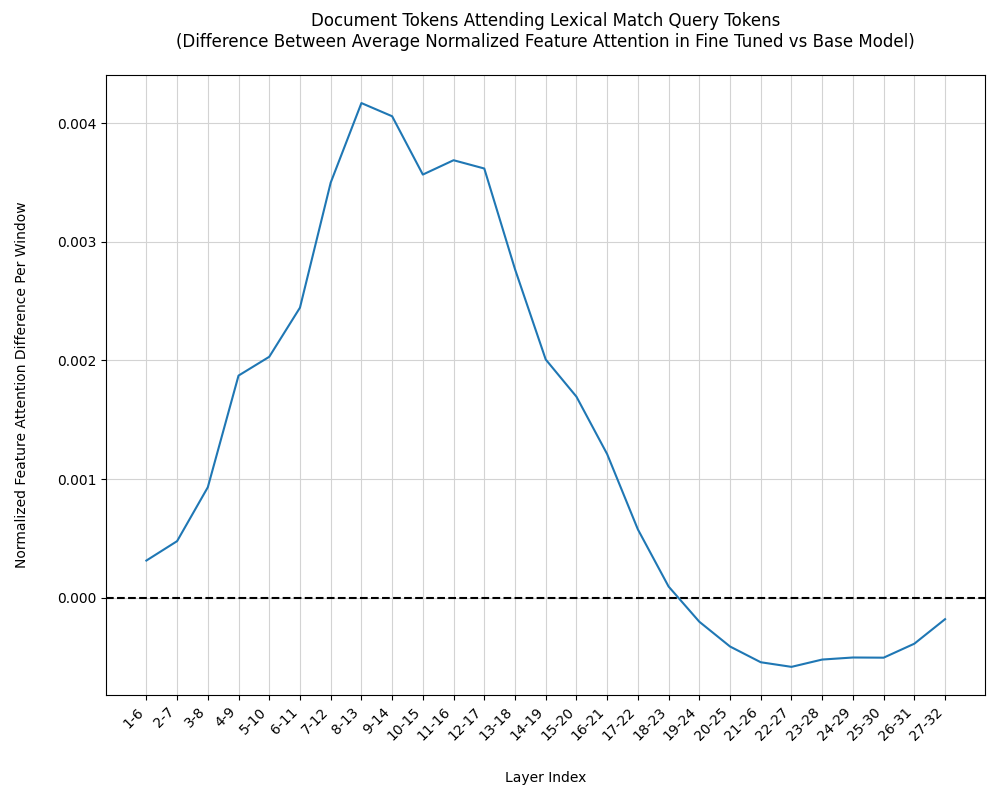}
    \caption{Change in normalized feature attention for document tokens attending lexical match query tokens, averaged over sliding windows of size 6, between the base and fine-tuned models.}
    \label{fig:composition2}
\end{figure*}

\begin{figure*}
    \centering
    \includegraphics[width=0.8\linewidth]{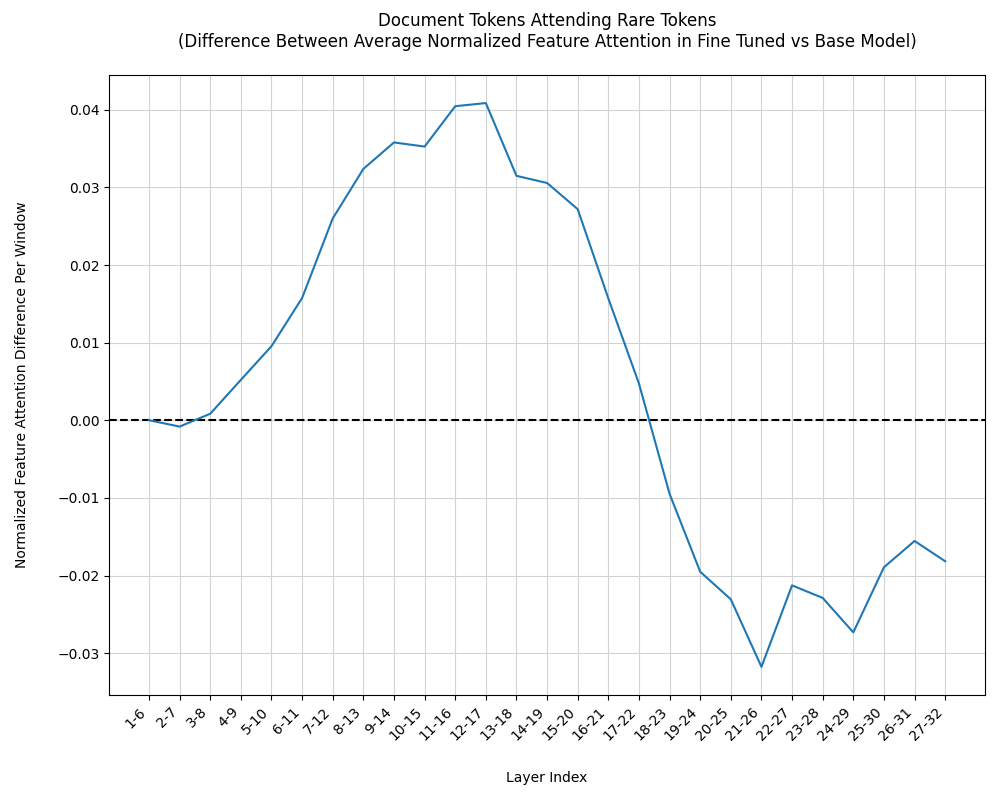}
    \caption{Change in normalized feature attention for document tokens attending rare tokens, averaged over sliding windows of size 6, between the base and fine-tuned models.}
    \label{fig:composition3}
\end{figure*}

Since there are many compositional features, we limit our view to features with a strong correlation and interpretable meaning.
Figures~\ref{fig:composition1}-\ref{fig:composition3} show the difference between normalized feature attention before and after fine-tuning, per window, for several compositional features of interest.

Table~\ref{tab:composition-spearman} lists the correlations between learned attention to these compositional features and the NDCG impact of ablating them, demonstrating several compositions are more indicative of fine-tuned relevance-oriented learning than the features they are made up of.
In particular, learned document-query attention between rare words and lexical matches is highly predictive of performance gains, suggesting that neural models likely rely on shared rare words across segments the same way many classical models do.

\clearpage

\clearpage

\begin{table*}[h]
\centering
\small
\begin{tabular}{lcc}
\toprule
\textbf{Feature} & \textbf{Keep} $\rho$ & \textbf{Omit} $\rho$ \\
\midrule
Rare Document Tokens Attending \\ Rare Lexical Match Query Tokens & 0.88 & $-$0.94 \\
\midrule
Document Tokens Attending \\ Lexical Match Query Tokens & 0.92 & $-$0.92 \\
\midrule
Document Tokens \\ Attending Rare Tokens & 0.92 & $-$0.88 \\
\bottomrule
\end{tabular}
\caption{Spearman correlation between the effect of per-window ablations on NDCG performance and per-window change in normalized feature attention after fine-tuning. Keep correlations are positive because windows that learn stronger feature attention also produce larger performance gains when kept in isolation; omit correlations are negative because those same windows produce larger performance drops when removed.}
\label{tab:composition-spearman}
\end{table*}

\clearpage

\section{Additional Window Size Ablations}
\label{app:window-sizes}

In addition to the 6-layer window ablations pictured in Figure~\ref{fig:omit-window} and Figure~\ref{fig:keep-window},
we investigate ablations on windows consisting of 3 and 4 layers.

\subsection{Omit Ablations}

\begin{figure}[H]
    \includegraphics[width=\linewidth]{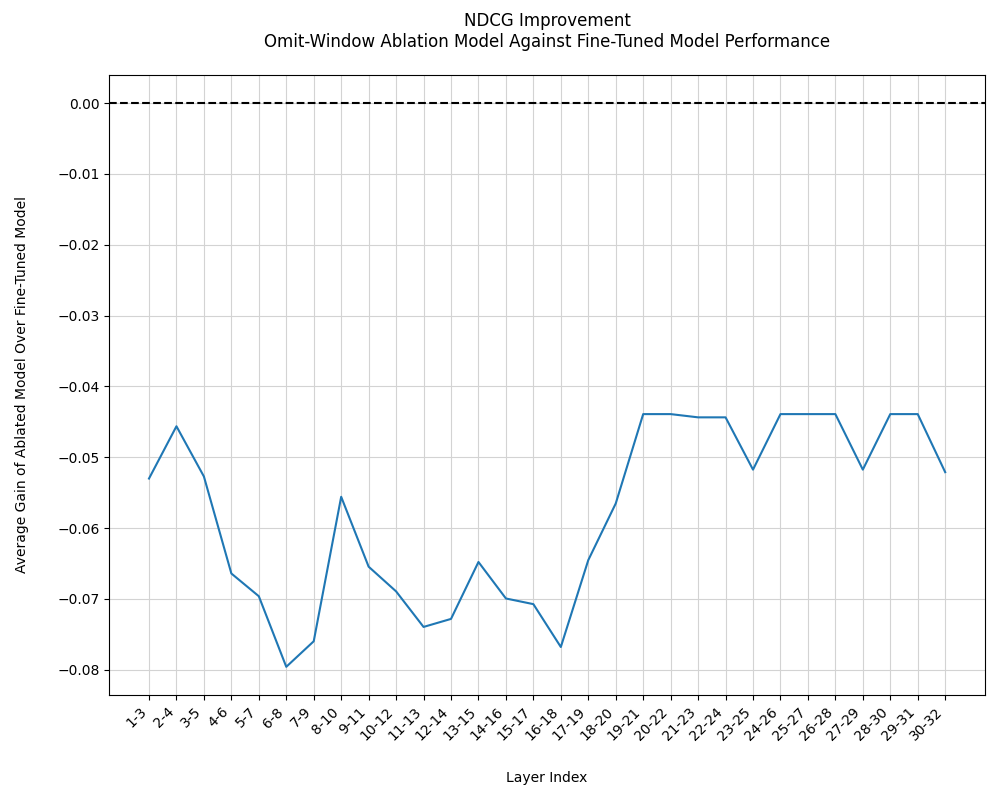}
    \caption{Omit-window results using window size 3.}
    \label{fig:omit-window-s3}
\end{figure}

\begin{figure}[H]
    \includegraphics[width=\linewidth]{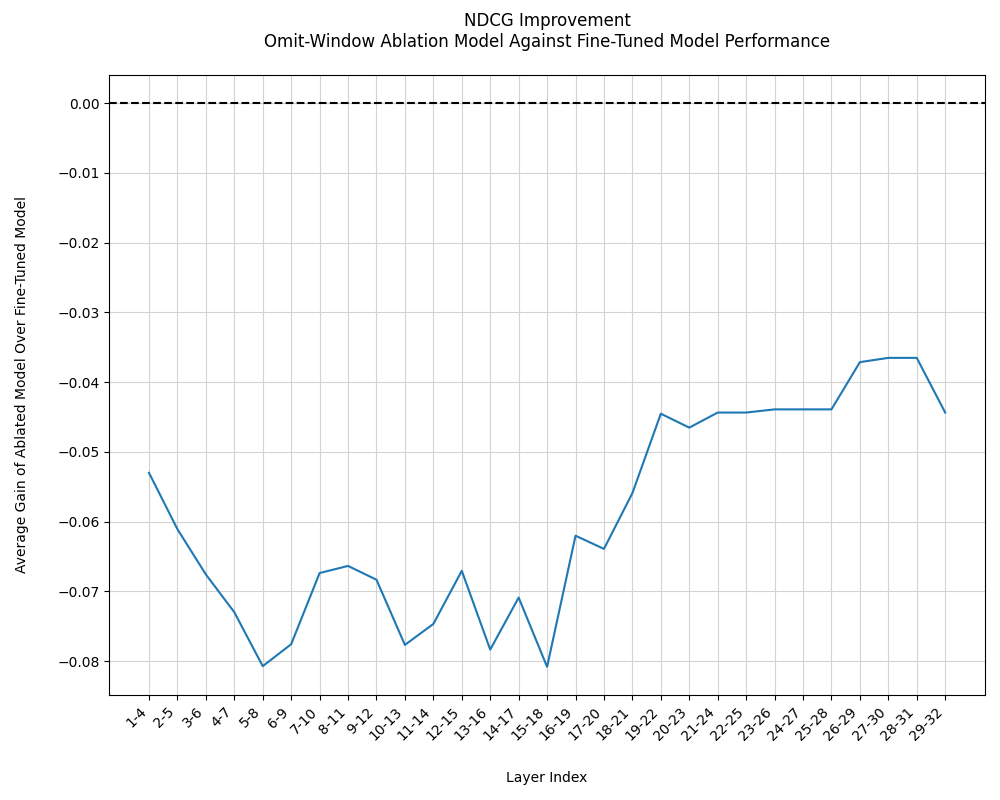}
    \caption{Omit-window results using window size 4.}
    \label{fig:omit-window-s4}
\end{figure}

Omit-window ablation results across the additional window sizes (Figures~\ref{fig:omit-window-s3}-\ref{fig:omit-window-s4}) yield similar trends to Figure~\ref{fig:omit-window}, with fine-tuning for layers 6--18 emerging as most necessary for performance.

\subsection{Keep Ablations}

\begin{figure}[H]
    \includegraphics[width=\linewidth]{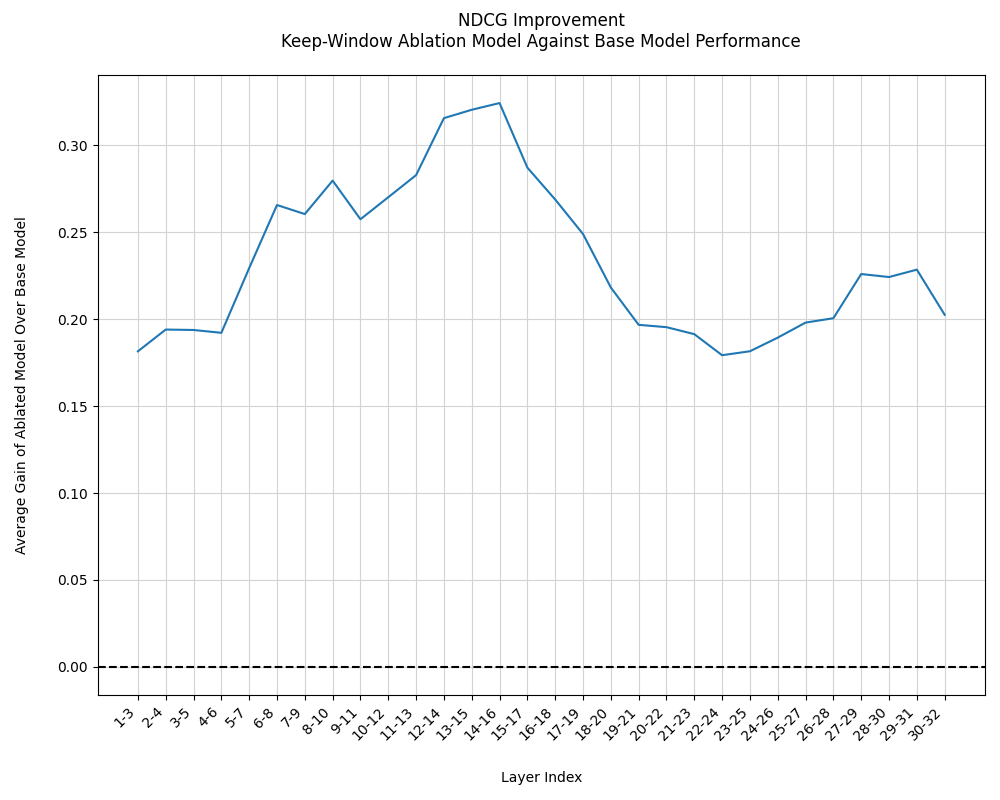}
    \caption{Keep-window results using window size 3.}
    \label{fig:keep-window-s3}
\end{figure}

\begin{figure}[H]
    \includegraphics[width=\linewidth]{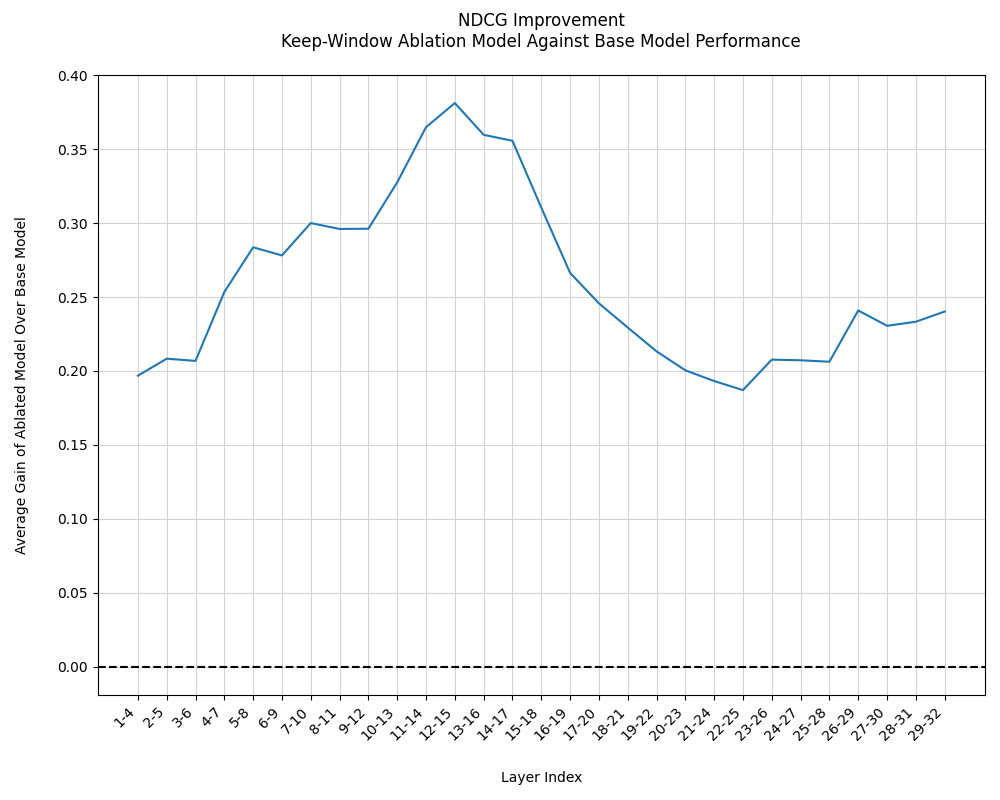}
    \caption{Keep-window results using window size 4.}
    \label{fig:keep-window-s4}
\end{figure}

Keep-window ablations for all the additional window sizes (Figures~\ref{fig:keep-window-s3}-\ref{fig:keep-window-s4}) demonstrate trends similar to Figure~\ref{fig:keep-window}, with layers 9--18 recovering optimal fine-tuning gains when kept in isolation.

\subsection{Attention Feature Correlation}

Table~\ref{tab:spearman-windows} reports Spearman correlations between learned feature attention and ablation performance for windows of sizes 3 and 4. Results are consistent across sizes, supporting the robustness of the trends reported in the main paper. Correlations are slightly lower for smaller window sizes, which we attribute to redundant components absorbing the effects of finer-grained ablations.

\begin{table}[H]
\centering
\small
\begin{tabular}{lcc}
\toprule
\textbf{Feature} & \textbf{Keep} $\rho$ & \textbf{Omit} $\rho$ \\
\midrule
\textit{Window Size 3} \vspace{0.25em} \\
Lexical Matching   & 0.06 & $-$0.39 \\
Rarity Sensitivity            & 0.79 & $-$0.80 \\
Document-Query Interaction      & 0.61 & $-$0.64 \\
\midrule
\textit{Window Size 4} \vspace{0.25em} \\
Lexical Matching   & 0.19 & $-$0.55 \\
Rarity Sensitivity            & 0.82 & $-$0.79 \\
Document-Query Interaction      & 0.63 & $-$0.69 \\
\bottomrule
\end{tabular}
\caption{Spearman correlation between the effect of per-window ablations on NDCG performance and per-window change in normalized feature attention after fine-tuning. Keep correlations are positive because windows that learn stronger feature attention also produce larger performance gains when kept in isolation; omit correlations are negative because those same windows produce larger performance drops when removed.}
\label{tab:spearman-windows}
\end{table}

\newpage

\section{Code and Reproducibility}

Our ablation and attention experiments use 7B-parameter models and require approximately 25 hours of compute on a single NVIDIA A100 GPU.
Anonymized code to reproduce all experiments is available with instructions \href{https://anonymous.4open.science/r/lora-attention-interpretability-782C}{here}.

\clearpage


\newpage


\end{document}